\documentclass[letterpaper]{article}
\usepackage[letterpaper, margin=1.5in]{geometry}

\usepackage{lineno,hyperref}
\usepackage{rotating}
\usepackage{multirow}
\usepackage{graphicx}
\usepackage{booktabs}
\usepackage{color}
\usepackage{sidecap}
\usepackage{wrapfig}
\usepackage[font=scriptsize]{subcaption}
\usepackage{tablefootnote}
\modulolinenumbers[5]
\usepackage{color}
\usepackage{enumitem}
\usepackage{authblk}
\usepackage{float}
\usepackage{tikz}
\newcommand{\tabitem}{~~\llap{\textbullet}~~}
\def\checkmark{\tikz\fill[scale=0.4](0,.35) -- (.25,0) -- (1,.7) -- (.25,.15) -- cycle;}

\begin{document}
\title{Visual Question Answering: \\Datasets, Algorithms, and Future Challenges}

\author[ ]{Kushal Kafle}
\author[ ]{Christopher Kanan \thanks{Corresponding author.}}
\affil[ ]{Chester F. Carlson Center for Imaging Science \protect\\ Rochester Institute of Technology, Rochester, NY, 14623, USA}
\affil[ ]{\texttt{{kk6055,kanan@rit.edu}}}
\date{}
\maketitle
\begin{abstract}
Visual Question Answering (VQA) is a recent problem in computer vision and natural language processing that has garnered a large amount of interest from the deep learning, computer vision, and natural language processing communities. In VQA, an algorithm needs to answer text-based questions about images. Since the release of the first VQA dataset in 2014, additional datasets have been released and many algorithms have been proposed. In this review, we critically examine the current state of VQA in terms of problem formulation, existing datasets, evaluation metrics, and algorithms. In particular, we discuss the limitations of current datasets with regard to their ability to properly train and assess VQA algorithms. We then exhaustively review existing algorithms for VQA. Finally, we discuss possible future directions for VQA and image understanding research.
\end{abstract}

\section{Introduction} \label{sec:intro}
Recent advancements in computer vision and deep learning research have enabled enormous progress in many computer vision tasks, such as image classification~\cite{Simonyan14,resnet}, object detection~\cite{redmon2015you,ren2015faster}, and activity recognition~\cite{lrcn2014,karpathy2014large,simonyan2014two}. Given enough data, deep convolutional neural networks (CNNs) rival the abilities of humans to do image classification~\cite{resnet}. With annotated datasets rapidly increasing in size thanks to crowd-sourcing, similar outcomes can be anticipated for other focused computer vision problems. However, these problems are narrow in scope and do not require holistic understanding of images. As humans, we can identify the objects in an image, understand the spatial positions of these objects, infer their attributes and relationships to each other, and also reason about the purpose of each object given the surrounding context. We can ask arbitrary questions about images and also communicate the information gleaned from them.

Until recently, developing a computer vision system that can answer arbitrary natural language questions about images has been thought to be an ambitious, but intractable, goal. However, since 2014, there has been enormous progress in developing systems with these abilities. Visual Question Answering (VQA) is a computer vision task where a system is given a text-based question about an image, and it must infer the answer. Questions can be arbitrary and they encompass many sub-problems in computer vision, e.g.,
\begin{itemize}[noitemsep]
\item Object recognition - What is in the image?
\item Object detection - Are there any cats in the image?
\item Attribute classification - What color is the cat?
\item Scene classification - Is it sunny?
\item Counting - How many cats are in the image?
\end{itemize}
Beyond these, there are many more complex questions that can be asked, such as questions about the spatial relationships among objects (What is between the cat and the sofa?) and common sense reasoning questions (Why is the the girl crying?). A robust VQA system must be capable of solving a wide range of classical computer vision tasks as well as needing the ability to reason about images. 

There are many potential applications for VQA. The most immediate is as an aid to blind and visually impaired individuals, enabling them to get information about images both on the web and in the real world. For example, as a blind user scrolls through their social media feed, a captioning system can describe the image and then the user could use VQA to query the image to get more insight about the scene. More generally, VQA could be used to improve human-computer interaction as a natural way to query visual content. A VQA system can also be used for image retrieval, without using image meta-data or tags. For example, to find all images taken in a rainy setting, we can simply ask `Is it raining?' to all images in the dataset. Beyond applications, VQA is an important basic research problem. Because a good VQA system must be able to solve many computer vision problems, it can be considered a component of a Turing Test for image understanding~\cite{geman2015visual,malinowski2014towards}.

A Visual Turing Test rigorously evaluates a computer vision system to assess whether it is capable of human-level semantic analysis of images~\cite{geman2015visual,malinowski2014towards}. Passing this test requires a system to be capable of many different visual tasks. VQA can be considered a kind of Visual Turing Test that also requires the ability to understand questions, but not necessarily more sophisticated natural language processing. If an algorithm performs as well as or better than humans on arbitrary questions about images, then arguably much of computer vision would be solved. But, this is only true if the benchmarks and evaluation tools are sufficient to make such bold claims. 

In this review, we discuss existing datasets and methods for VQA. 
We place particular emphasis on exploring whether current VQA benchmarks are suitable for evaluating whether a system is capable of robust image understanding. In Section \ref{sec:vqa-vs}, we compare VQA with other computer vision tasks, some of which also require the integration of vision and language (e.g., image captioning). Then, in Section \ref{sec:datasets}, we describe currently available datasets for VQA with an emphasis on their strengths and weaknesses. We discuss how biases in some of these datasets severely limit their ability to assess algorithms. In Section~\ref{sec:evaluation}, we discuss the evaluation metrics used for VQA. Then, we review existing algorithms for VQA and analyze their efficacy in Section~\ref{sec:previous}. Finally, we discuss possible future developments in VQA and open questions. 

\begin{figure*}[t]
	\centering \par\medskip
    \captionsetup[subfigure]{labelformat=empty,font=scriptsize}
	\begin{subfigure}[t]{0.3\textwidth}
		\includegraphics[width=\textwidth]{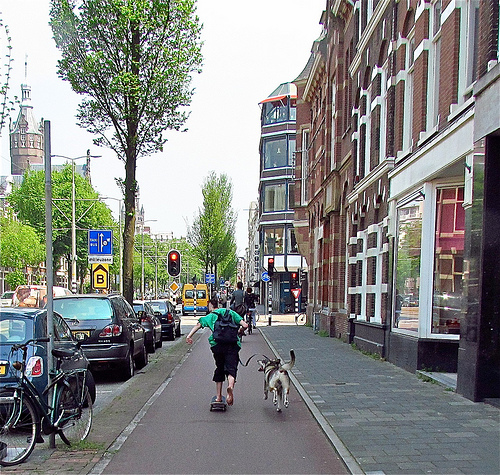}
	\end{subfigure}
    \hfill
	\begin{subfigure}[t]{0.3\textwidth}
		\includegraphics[width=\textwidth]{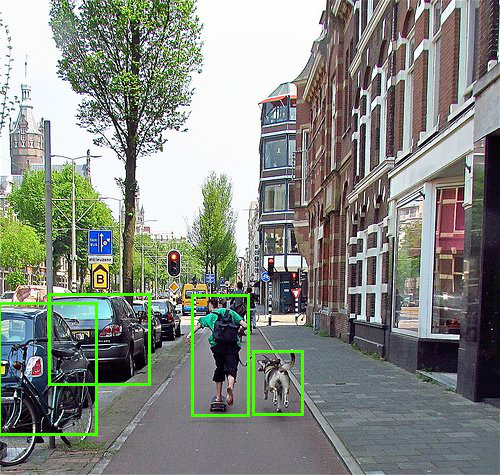}
		\end{subfigure}
        \hfill
        	\begin{subfigure}[t]{0.3\textwidth}
		\includegraphics[width=\textwidth]{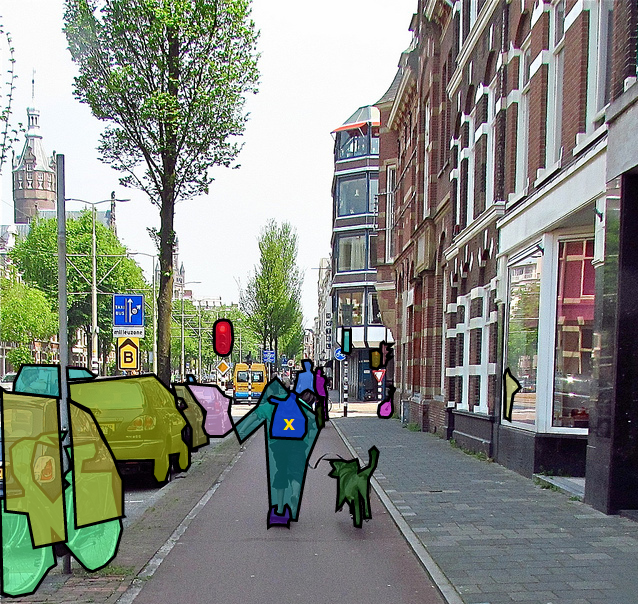}
		\end{subfigure}
        \hfill
		 \caption{ Object detection, semantic segmentation, and image captioning compared to VQA. The middle figure shows the ideal output of a typical object detection system, and the right figure shows the semantic segmentation map from the COCO dataset~\cite{lin2014microsoft}. Both tasks lack the ability to provide contextual information about the objects. The captions for this COCO image range from very generic descriptions of the scene, e.g., \textit{A busy town sidewalk next to street parking and intersections.}, to very focused discussion of a single activity without qualifying the overall scene, e.g., \textit{A woman jogging with a dog on a leash.} Both are acceptable captions, but significantly more information can be extracted with VQA. For the COCO-VQA dataset, the questions asked about this image are \textit{What kind of shoes is the skater wearing?}, \textit{Urban or suburban?}, and \textit{What animal is there?}}
    \label{fig:differenttasks}
\end{figure*}

\section{Vision and Language Tasks Related to VQA} \label{sec:vqa-vs}
The overarching goal of VQA is to extract question-relevant semantic information from the images, which ranges from the detection of minute details to the inference of abstract scene attributes for the whole image, based on the question. While many computer vision problems involve extracting information from the images, they are limited in scope and generality compared to VQA. Object recognition, activity recognition, and scene classification can all be posed as image classification tasks, with today's best methods doing this using CNNs trained to classify images into particular semantic categories. The most successful of these is object recognition, where algorithms now rival humans in accuracy~\cite{resnet}. But, object recognition requires only classifying the dominant object in an image without knowledge of its spatial position or its role within the larger scene. Object detection involves the localization of specific semantic concepts (e.g., cars or people) by placing a bounding box around each instance of the object in an image. The best object detection methods all use deep CNNs~\cite{szegedy2013deep,ren2015faster,redmon2015you}. Semantic segmentation takes the task of localization a step further by classifying each pixel as belonging to a particular semantic class~\cite{long2015fully,noh2015learning}. Instance segmentation further builds upon localization by differentiating between separate instances of the same semantic class~\cite{silberman2014instance,zhang2015monocular,zhang2015instance}.

While semantic and instance segmentation are important computer vision problems that generalize object detection and recognition, they are not sufficient for holistic scene understanding. One of the major problems they face is label ambiguity. For example, in Figure~\ref{fig:differenttasks}, the assigned semantic label for the position of the yellow cross can be `bag', `black,' or `person.' The label depends on the task. Moreover, these approaches alone have no understanding of the role of an object within a larger context. In this example, labeling a pixel as `bag' does not inform us about whether it is being carried by the person, and labeling a pixel as `person' does not tell us if the person is sitting, running, or skateboarding. This is in contrast with VQA, where a system is required to answer arbitrary questions about images, which may require reasoning about the relationships of objects with each other and the overall scene. The appropriate label is specified by the question.

Besides VQA, there is a significant amount of recent work that combines vision with language. One of the most studied is image captioning \cite{karpathy2014deep,lrcn2014,mao2014deep,vinyals2014show,xu2015show}, in which an algorithm's goal is to produce a natural language description of a given image. Image captioning is a very broad task that potentially involves describing complex attributes and object relationships to provide a detailed description of an image.

However, there are several problems with the visual captioning task, with evaluation of captions being a particular challenge. The ideal method is evaluation by human judges, but this is slow and expensive. For this reason, multiple automatic evaluation schemes have been proposed. The most widely used caption evaluation schemes are BLEU~\cite{papineni2002bleu}, ROUGE~\cite{lin2004rouge}, METEOR~\cite{banerjee2005meteor}, and CIDEr~\cite{vedantam2015cider}. With the exception of CIDEr, which was developed specifically for scoring image descriptions, all caption evaluation metrics were originally developed for machine translation evaluation. Each of these metrics has limitations. BLEU, the most widely used metric, is known to have the same score for large variations in sentence structure with largely varying semantic content~\cite{callison2006re}. For captions generated in \cite{fang2014captions}, BLEU scores ranked machine generated captions above human captions. However, when human judges were used to judge the same captions, only 23.3\% of the judges ranked the captions to be of equal or better quality than human captions. While other evaluation metrics, especially CIDEr and METEOR, show more robustness in terms of agreement with human judges, they still often rank automatically generated captions higher than human captions~\cite{bernardi2016automatic}. 

One reason why evaluating captions is challenging is that a given image can have many valid captions, with some being very specific and others generic in nature (see Figure~\ref{fig:differenttasks}). However, captioning systems that produce  generic captions that only superficially describe an image's content are often ranked high by the evaluation metrics. Generic captions such as `A person is walking down a street' or `Several cars are parked on the side of the road' that can be applicable to a large number of images are often ranked highly by evaluation schemes and human judges. In fact, a simple system that returns the caption of the training image with the most similar visual features using nearest neighbor yields relatively high scores using automatic evaluation metrics~\cite{devlin2015exploring}. 

Dense image captioning (DenseCap) avoids the generic caption problem by annotating an image densely with short visual descriptions pertaining to small, but salient, image regions~\cite{johnson2015densecap}. For example, a DenseCap system may output `a man wearing black shirt,' 'large green trees,' and `roof of a building,' with each description accompanied by a bounding box. A system may generate a large number of these descriptions for rich scenes. Although many of these descriptions are short, it is still difficult to automatically assess their quality. DenseCap can also omit important relationships between the objects in the scene by only producing isolated descriptions for each regions. Captioning and DenseCap are also task agnostic and a system is not required to perform exhaustive image understanding.

In conclusion, a captioning system is at liberty to arbitrarily choose the level of granularity of its image analysis which is in contrast to VQA, where the level of granularity is specified by the nature of the question asked. For example, `What season is this?' will require understanding the entire scene, but `What is the color of dog standing behind the girl with white dress?' would require attention to specific details of the scene. Moreover, many kinds of questions have specific and unambiguous answers, making VQA  more amenable to automated evaluation metric than captioning. Ambiguity may still exist for some question types (see Section \ref{sec:evaluation}), but for many questions the answer produced by a VQA algorithm can be evaluated with one-to-one matching with the ground truth answer.
\section{Datasets for VQA} \label{sec:datasets}
\begin{sidewaystable}
\centering
\caption{Statistics for VQA datasets using either open-ended (OE) or multiple-choice (MC) evaluation schemes.}
\label{table:datasets}
\begin{tabular}{@{}lrrrrrr@{}}
\toprule
                       & DAQUAR~\cite{malinowski2014multi}              & COCO-QA~\cite{ren2015image}    & COCO-VQA~\cite{antol2015vqa} & FM-IQA~\cite{gao2015you}~\tablefootnote{We were unable to retrieve the English version of the dataset from provided download link.} & Visual7W~\cite{visual7w} & Visual genome~\cite{krishnavisualgenome} \\ \midrule
Total Images           & 1,449                & 123,287     & 204,721   & 120,360 & 47,300    & 108,000        \\
QA Pairs               & 12,468               & 117,684     & 614,163   & 250,569 & 327,939   & 1,773,258     \\
Distinct Answers       & 968                 & 430        & 105,969   & N/A    & 65,161    & 207,675        \\
\% covered by top-1000 & 100               & 100      & 82.8   & N/A    & 56.29  & 60.8        \\
\% covered by top-10   & 25.04             & 19.71    & 51.13  & N/A    & 17.13  & 13.07       \\
Human Accuracy & 50.2& N/A & 83.3&N/A& 96.6 &N/A\\
Longest Question       & 25 words                 & 24 words        & 32 words      & N/A    & 24 words      & 26 words      \\
Longest Answer         & 7 (list of 1 words) & 1 word     & 17 words & N/A    & 20 words & 24 words      \\
Avg. Answer Length & 1.2 words & 1.0 words & 1.1 words & N/A & 2.0 words & 1.8 words\\
Image Source           & NYUDv2              & COCO       & COCO    & COCO   & COCO     & COCO, YFCC    \\
Annotation             & Manual+Auto    & Auto & Manual    & Manual  & Manual    & Manual         \\
Evaluation Type             & OE    & OE & MC or OE    & OE  & MC or OE    & OE         \\
Question Types         & 3                   & 4          & -      & -    & -        & -           \\ \bottomrule
\end{tabular}
\end{sidewaystable}

Beginning in 2014, five major datasets for VQA have been publicly released. These datasets enable VQA systems to be trained and evaluated. As of this article, the main datasets for VQA are DAQUAR~\cite{malinowski2014multi}, COCO-QA~\cite{ren2015image}, The VQA Dataset~\cite{antol2015vqa}, FM-IQA~\cite{gao2015you}, Visual7W~\cite{visual7w}, and Visual Genome~\cite{krishnavisualgenome}. With exception of DAQUAR, all of the datasets include images from the Microsoft Common Objects in Context (COCO) dataset~\cite{lin2014microsoft}, which consists of 328,000 images, 91 common object categories with over 2 million labeled instances, and an average of 5 captions per image. Visual Genome and Visual7W use images from Flickr100M in addition to the COCO images.  A portion of The VQA Dataset contains synthetic cartoon imagery, which we will refer to as SYNTH-VQA. Consistent with other papers~\cite{kafle2016,WuWSHD15,zhou2015}, the rest of The VQA Dataset will be referred as COCO-VQA, since it contains images from the COCO image dataset. Table \ref{table:datasets} contains statistics for each of these datasets.

An ideal VQA dataset needs to be sufficiently large to capture the variability  within questions, images, and concepts that occur in real world scenarios. It should also have a fair evaluation scheme that is difficult to `game' and doing well on it indicates that an algorithm can answer a large variety of question types about images that have definitive answers.  If a dataset contains easily exploitable biases in the distribution of the questions or answers, it may be possible for an algorithm to perform well on the dataset without really solving the VQA problem. 

In the following subsections, we critically review the available datasets. We describe how the datasets were created and discuss their limitations.

\subsection{DAQUAR}
The DAtaset for QUestion Answering on Real-world images (DAQUAR) \cite{malinowski2014multi} was the first major VQA dataset to be released. It is one of the smallest VQA datasets. It consists of 6795 training and 5673 testing QA pairs based on images from the NYU-DepthV2 Dataset~\cite{Silberman}. The dataset is also available in an even smaller configuration consisting of only 37 object categories, known as DAQUAR-37. DAQUAR-37 consists of only 3825 training QA pairs and 297 testing QA pairs. In~\cite{malinowskiRF15}, additional ground truth answers were collected for DAQUAR to create an alternative evaluation metric. This variant of DAQUAR is called DAQUAR-consensus, named after the evaluation metric. While DAQUAR was a pioneering dataset for VQA, it is too small to successfully train and evaluate more complex models. Apart from the small size, DAQUAR contains exclusively indoor scenes, which constrains the variety of questions available.  The images tend to have significant clutter and in some cases extreme lighting conditions (see Figure~\ref{fig:sample_images}). This makes many questions difficult to answer, and even humans are only able to achieve 50.2\% accuracy on the full dataset.

\subsection{COCO-QA}
\begin{figure*}[t]
	\centering \par\medskip
    \captionsetup[subfigure]{labelformat=empty,font=scriptsize}
	\begin{subfigure}[t]{0.49\textwidth}
		\includegraphics[width=\textwidth]{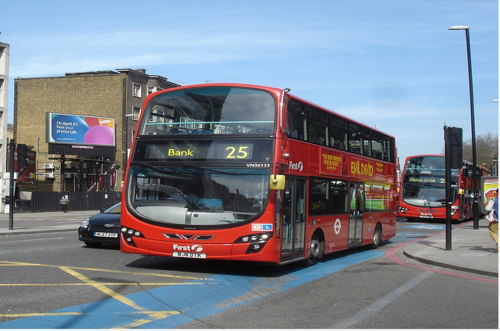}
		 \caption{ COCO-QA: \color{black} What does an intersection show on one side and two double-decker buses and a third vehicle,?\\ Ground Truth:  \color{black}Building}
	\end{subfigure}
    \hfill
	\begin{subfigure}[t]{0.43\textwidth}
		\includegraphics[width=\textwidth]{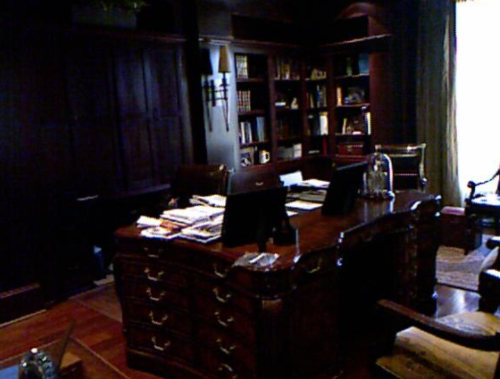}
		\caption{DAQUAR: \color{black}What is behind the computer in the corner of the table?\\ Ground Truth:  \color{black}papers}
		\end{subfigure}
        \hfill
        \caption{Sample images from DAQUAR and the COCO-QA datasets and the corresponding QA pairs. A significant number of COCO-QA questions have grammatical errors and are nonsensical, whereas DAQUAR images are often marred with clutter and low resolution images.}
    \label{fig:sample_images}
\end{figure*}

In COCO-QA \cite{ren2015image}, QA pairs are created for images using an Natural Language Processing (NLP) algorithm that derives them from the COCO image captions. For example, using the image caption \texttt{A boy is playing Frisbee}, it is possible to create the question \texttt{What is the boy playing?} with \texttt{frisbee} as the answer. COCO-QA contains 78,736 training and 38,948 testing QA pairs. Most questions ask about the object in the image (69.84\%), with the other questions being about color (16.59\%), counting (7.47\%) and location (6.10\%). All of the questions have a single word answer, and there are only 435 unique answers. These constraints on the answers makes evaluation relatively straightforward.

The biggest shortcoming of COCO-QA is due to flaws in the NLP algorithm that was used to generate the QA pairs.  Longer sentences are broken into smaller chunks for ease of processing, but in many of these cases the algorithm does not cope well with the presence of clauses and grammatical variations in sentence formation. This results in awkwardly phrased questions, with many containing grammatical errors, and others being completely unintelligible (see Figure  \ref{fig:sample_images}). The other major shortcoming is that it only has four kinds of questions, and these are limited to the kinds of things described in COCO's captions. 

\subsection{The VQA Dataset}
The VQA Dataset~\cite{antol2015vqa} consists of both real images from COCO and abstract cartoon images. Most work on this dataset has focused solely on the portion containing real world imagery from COCO, which we refer to as COCO-VQA. We refer to the synthetic portion of the dataset as SYNTH-VQA.

COCO-VQA consists of three questions per image, with ten answers per question. Amazon Mechanical Turk (AMT) workers were employed to generate questions for each image by being asked to `Stump a smart robot,' and a separate pool of workers were hired to generate the answers to the questions. Compared to other VQA datasets, COCO-VQA consists of a relatively large number of questions (614,163 total, with 248,349 for training, 121,512 for validation, and 244,302 for testing). Each of the questions is then answered by 10 independent annotators. The multiple answers per question are used in the consensus-based evaluation metric for the dataset, which is discussed in Section \ref{sec:evaluation}.

SYNTH-VQA consists of 50,000 synthetic scenes that depict cartoon images in different simulated scenarios. Scenes are made from over 100 different objects, 30 different animal models, and 20 human cartoon models. The human models are the same as those used in \cite{antol2014zero}, and they contain deformable limbs and eight different facial expressions. The models also span different age, gender, and races to provide variation in appearance. SYNTH-VQA has 150,000 QA pairs with 3 questions per scene and 10 ground truth answers per question. By using synthetic images, it becomes possible to create a more varied and balanced dataset. Natural image datasets tend to have more consistent context and biases, e.g., a street scene is more likely to have picture of a dog than a zebra. Using synthetic images, these biases can be reduced. Yin and Yang \cite{Zhang_2016_CVPR} is a dataset built on top of SYNTH-VQA that tried to eliminate biases in the answers people have to questions. We further discuss Yin and Yang in Section~\ref{subsec:bias}.

Both SYNTH-VQA and COCO-VQA come in both open-ended and multiple-choice formats. The multiple-choice format contains all the same QA pairs, but it also contains 18 different choices that are comprised of
\begin{itemize}[noitemsep]
\item \textbf{The Correct Answer}, which is the most frequent answer given by the ten annotators.
\item \textbf{Plausible Answers}, which are three answers collected from annotators without looking at the image.
\item \textbf{Popular Answers}, which are the top ten most popular answers in the dataset. 
\item \textbf{Random Answers}, which are randomly selected correct answers for other questions.
\end{itemize}

\begin{figure*}[t]
	\centering \par\medskip
    \captionsetup[subfigure]{font=scriptsize}
	\begin{subfigure}[t]{0.3\textwidth}
		\includegraphics[width=\textwidth]{images/vqa1.png}
		 \caption{Q: Would you like to fly in that? GT: yes (4x), no (6x). The VQA Dataset contains subjective questions that are prone to cause disagreement between annotators and also clearly lack a single objectively correct answer.}
	\end{subfigure}
    \hfill
	\begin{subfigure}[t]{0.3\textwidth}
		\includegraphics[width=\textwidth]{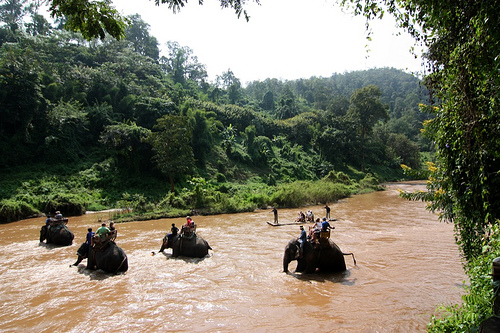}
		\caption{Q: What color are the trees? GT: green. There are 73 total questions in the dataset asking this question. For 70 of those questions, the majority answer is green. Such questions can be often answered without information from the image.}
		\end{subfigure}
        \hfill
        \begin{subfigure}[t]{0.34\textwidth}
		\includegraphics[width=\textwidth]{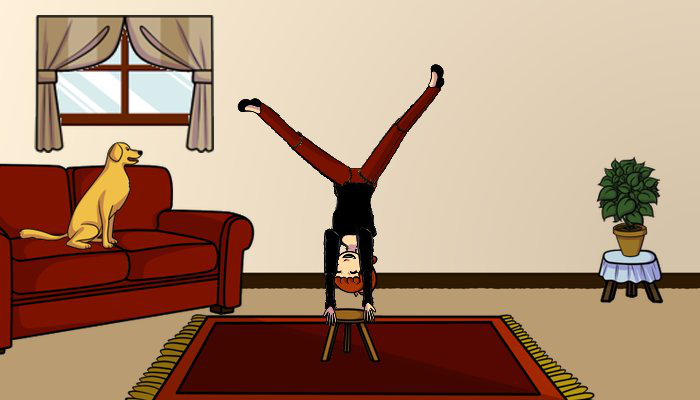}
		\caption{Q: Why would you say this woman is strong? GT: yes (5x), can lift up on arms, headstand, handstand, can stand on her head, she is standing upside down on stool. Questions seeking descriptive or explanatory answers can pose significant difficulty in evaluation.}
         \label{fig:vqa3}
		\end{subfigure}
        \hfill
        \caption{Open ended QA pairs from The VQA Dataset for both real and abstract images.}
    \label{fig:vqa}
\end{figure*}
   
Due to diversity and size of the dataset, COCO-VQA has been widely used to evaluate algorithms. However, there are multiple problems with the dataset. COCO-VQA has a large variety of questions, but many of them can be accurately answered without using the image due to language biases.  Relatively simple image-blind algorithms have achieved 49.6\% accuracy on COCO-VQA using the question alone~\cite{kafle2016}. The dataset also contains many subjective, opinion-seeking questions that do not have a single objective answer (see Figure \ref{fig:vqa}). Similarly, many questions seek explanations or verbose descriptions. An example of this is given in Figure \ref{fig:vqa3}, which also shows unreliability of human annotators as the most popular answer is `yes' which is completely wrong for the given question. These complications are reflected by inter-human agreement on this dataset, which is about 83\%. Several other practical issues also arise out of the dataset's biases. For example, `yes/no' answers span about 38\% of all questions, and almost 59\% of them are answered with `yes.' Combined with the evaluation metric used with COCO-VQA (see Section \ref{sec:evaluation}), these biases can make it difficult to assess whether an algorithm is truly solving the VQA problem using solely this dataset. We discuss this further in Section \ref{sec:evaluation}.

\subsection{FM-IQA}
The Freestyle Multilingual Image Question Answering (FM-IQA) dataset is another dataset based on COCO~\cite{gao2015you}. It contains human generated answers and questions. The dataset was originally collected in Chinese, but English translations have been made available. Unlike COCO-QA and DAQUAR, this dataset also allowed for answers to be full sentences. This makes automatic evaluation with common metrics intractable. For this reason, the authors suggested using human judges for evaluation, where the judges are tasked with deciding whether or not the answer is provided by a  human or not as well as assessing the quality of an answer on a scale of 0--2. This approach is impractical for most research groups and makes developing algorithms difficult. We further discuss the importance of automatic evaluation metrics in Section \ref{sec:evaluation}.
\subsection{Visual Genome}
Visual Genome~\cite{krishnavisualgenome} consists of 108,249 images that occur in both YFCC100M \cite{thomee2016yfcc100m} and COCO images. It contains 1.7 million QA pairs for images, with an average of 17 QA pairs per image. As of this article, Visual Genome is the largest VQA dataset. Because it was only recently introduced, no methods have been evaluated on it beyond the baselines established by the authors.

\begin{figure*}[t]
	\centering \par\medskip
    \captionsetup[subfigure]{font=scriptsize}
	\begin{subfigure}[t]{0.45\textwidth}
		\includegraphics[width=\textwidth]{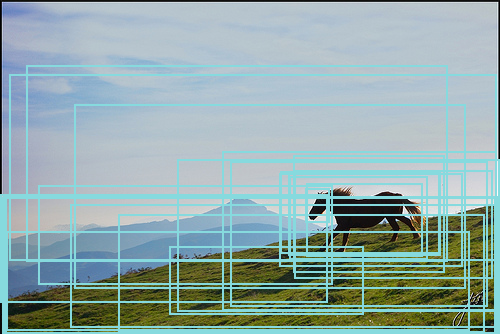}
		 \caption{Example image from the Visual Genome dataset along with annotated image regions. This figure is taken from \cite{krishnavisualgenome}. \\ Free form QA: What does the sky look like? \\ Region based QA: What color is the horse? }
          \label{subfig:visualgenome}
	\end{subfigure}
    \hfill
	\begin{subfigure}[t]{0.45\textwidth}
		\includegraphics[width=\textwidth]{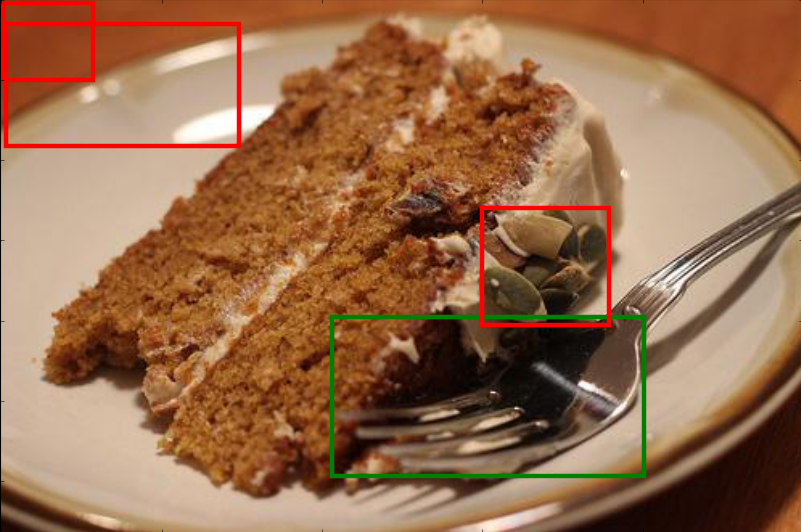}
		\caption{Example of the pointing QA task in Visual7W~\cite{visual7w}. The bounding boxes are the given choices. Correct answer is shown in green \\ Q: Which object can you stab food with?}\
        \label{subfig:visual7w}
		\end{subfigure}
        \caption{Visual7W is a subset of Visual Genome. Apart from the pointing task, all of the questions in Visual7W are sourced from Visual Genome data. Visual Genome, however, includes more than just QA pairs, such as region annotations.}
\end{figure*}

Visual Genome consists of six types of `W' questions: \texttt{What, Where, How, When, Who, and Why}. Two distinct modes of data collection were used to make the dataset. In the free-form method, annotators were free to ask any question about an image. However, when asking free-form questions, human annotators tend to ask similar questions about an image's holistic content, e.g., asking `How many horses are there?' or `Is it sunny?' This can  promote bias in the kinds of questions asked. The creators of Visual Genome combated this by also prompting workers to ask questions about specific image regions. When using this region-specific method, a worker might be prompted to ask a question about a region of an image containing a fire hydrant. Region-specific question prompting was made possible using Visual Genome's descriptive bounding-box annotations. An example of region bounding boxes and QA pairs from Visual Genome are shown in Figure \ref{subfig:visualgenome}.

\begin{figure*}[!t]
	\centering
		\includegraphics[width=\textwidth]{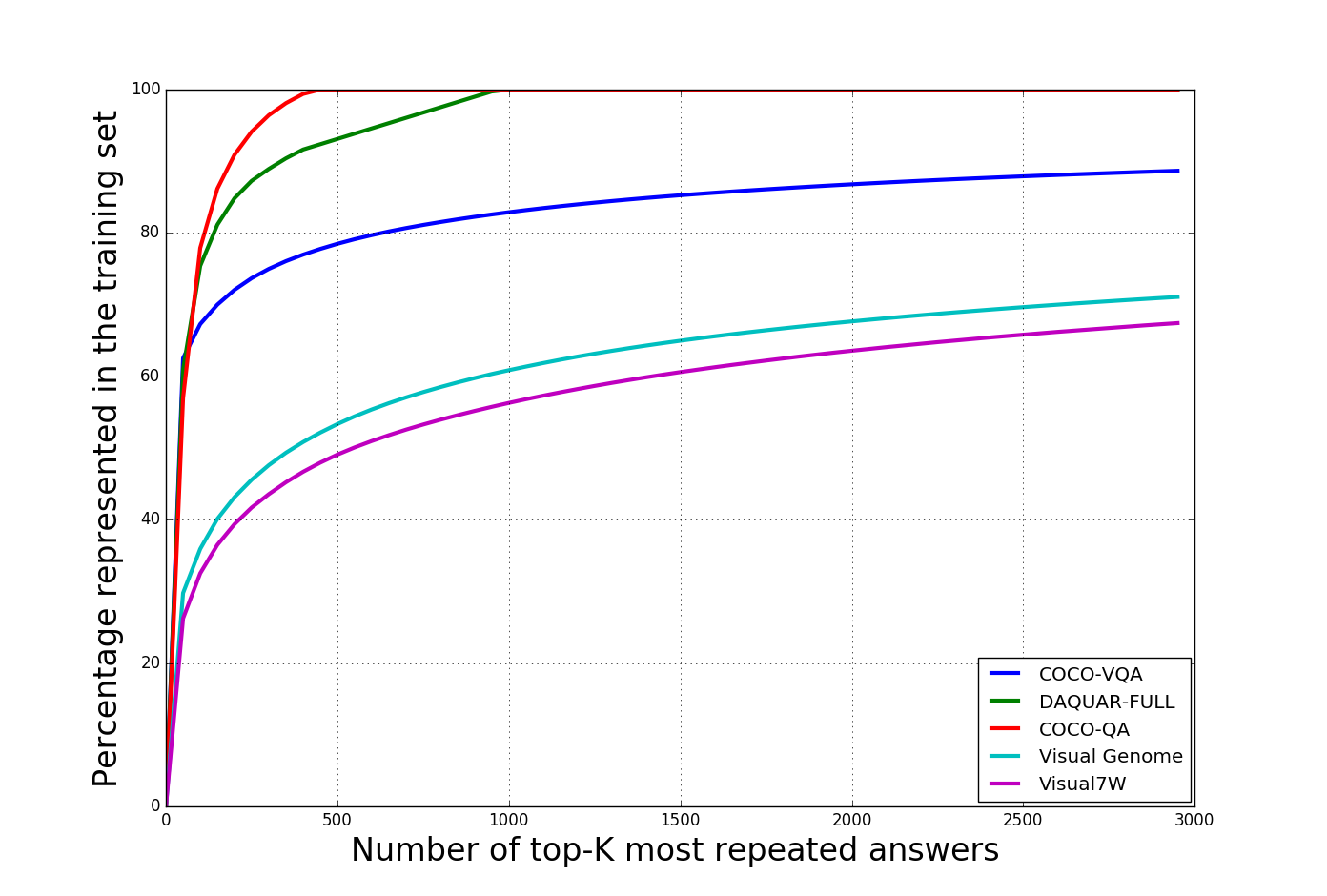}        \hfill
    \caption{This graph shows the long-tailed nature of answer distributions in newer VQA datasets. For example, choosing the 500 most repeated answers in the training set would cover a 100\% of all possible answers in COCO-QA but less than 50\% in the Visual Genome dataset. For classification based frameworks, this translates to training a model with more output classes.}
    \label{fig:dataset_stats}
\end{figure*}

Visual Genome has much greater answer diversity compared to other datasets, which is shown in Figure \ref{fig:dataset_stats}. The 1000 answers that occur most frequently in Visual Genome only cover 65\% of all answers in the dataset, whereas they cover 82\% for COCO-VQA and 100\% for DAQUAR and COCO-QA. Visual Genome's long-tailed distribution is also observed in the length of the answers. Only 57\% of answers are single words, compared to 88\% of answers in COCO-VQA, 100\% of answers in COCO-QA, and 90\% of answers in DAQUAR. This diversity in answers makes open-ended evaluation significantly more challenging. Moreover, since the categories themselves are required to strictly belong to one of the six `W' types, the diversity in answer may at times artificially stem simply from variations in phrasing which could be eliminated by prompting the annotators to choose more concise answers. For example, \textit{Where are the cars parked?} can be answered with `on the street' or more concisely with `street.' 

Visual Genome has no binary (yes/no) questions. The dataset creators argue that this will encourage using more complex questions. This is in contrast to The VQA Dataset, where `yes' and `no' are the more frequent answers in the dataset. We discuss this issue further in Section~\ref{sec:binary-questions}.

\subsection{Visual7W}
The Visual7W dataset is a subset of Visual Genome. Visual7W contains 47,300 images from Visual Genome that are also present in COCO. Visual7W is named after the seven categories of questions it contains: \texttt{What, Where, How, When, Who, Why, and Which}. The dataset consists of two distinct types of questions. The `telling' questions are identical to Visual Genome questions, and the answer is text-based. The `pointing' questions are the ones that begin with `Which,' and for these questions the algorithm has to select the correct bounding box among alternatives. An example pointing question is shown in Figure \ref{subfig:visual7w}.

Visual7W uses a multiple-choice answer framework as the standard evaluation, with four possible answers being made available to an algorithm during evaluation. To make the task challenging, the multiple-choices consist of answers that are plausible for the given question. Plausible answers are collected by prompting annotators to answer the question without seeing the image. For pointing questions, the multiple-choice options are four plausible bounding boxes surrounding the likely answer. Like Visual Genome, the dataset does not contain any binary questions. 
\subsection{SHAPES}
\begin{figure}
\centering 
\includegraphics[width=0.6\textwidth]{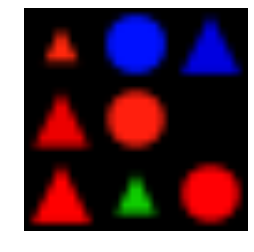}
        \caption{Example image from the SHAPES dataset. Questions in the SHAPES dataset \cite{AndreasRDK15} include counting (How many triangles are there?), spatial reasoning (Is there a red shape above a circle?), and  inference (Is there a blue shape red?)}
\label{fig:shapes}
    \end{figure}

While the other VQA datasets contain either real or synthetic scenes, the SHAPES dataset~\cite{AndreasRDK15} consists of shapes of varying arrangements, types, and colors. Questions are about the attributes, relationships, and positions of the shapes. This approach enables the creation of a vast amount of data, free of many of the biases that plague other datasets to varying degrees.

SHAPES consists of 244 unique questions, with every question asked about each of the 64 images in the dataset. Unlike other datasets, this means it is completely balanced and free of bias. All questions are binary, with yes/no answers. Many of the questions require positional reasoning about the layout and properties of the shapes. While, SHAPES cannot be a substitute for using real-world imagery, the idea behind it is extremely valuable. An algorithm that cannot perform well on SHAPES, but performs well on other VQA datasets may indicate that it is only capable of analyzing images in a limited manner. 

\section{Evaluation Metrics for VQA} \label{sec:evaluation}
\begin{figure}[t]
\centering
		\includegraphics[width=0.6\textwidth]{images/vqa_disagreement.png}
        \caption{Simple questions can also evoke diverse answers from annotators in COCO-VQA. Q: Where is the dog? A: 1) eating out of his bowl; 2) on floor; 3) feeding station; 4) by his food; 5) inside; 6) on floor eating out of his dish; 7) floor; 8) in front of gray bowl, to right of trash can; 9) near food bowl; 10) on floor}
        \label{fig:vqa_disagreement}
    \end{figure}

VQA has been posed as either an open-ended task, in which an algorithm generates a string to answer a question, or as a multiple-choice question where it picks among choices. For multiple-choice, simple accuracy is often used to evaluate, with an algorithm getting an answer right if it makes the correct choice. For open-ended VQA, simple accuracy can also be used. In this case, an algorithm's predicted answer string must exactly match the ground truth answer. However, accuracy can be too stringent because some errors are much worse than others. For example, if the question was `What animals are in the photo?' and a system outputs `dog' instead of the correct label `dogs,' it is penalized just as strongly as it would be if it output `zebra.'  Questions may also have multiple correct answers, e.g., `What is in the tree?' might have `bald eagle' listed as the correct ground truth answer, so a system that outputs `eagle' or `bird' would be penalized just as much as if it had output `yes' as the answer. Due to these issues, several alternatives to exact accuracy have been proposed for evaluating open-ended VQA algorithms.

     

\begin{table}[t]
\centering
\caption{Comparison of different evaluation metrics proposed for VQA.}
\label{table:metrics}
\begin{tabular}{@{}cll@{}}
\toprule
\multicolumn{1}{l}{}                                                 & \multicolumn{1}{c}{\textbf{Pros}}                                                                                                                                 & \multicolumn{1}{c}{\textbf{Cons}}                                                                                                                                                                              \\ \midrule

\textbf{\begin{tabular}[c]{@{}c@{}}Simple \\ Accuracy\end{tabular}}  & \begin{tabular}[c]{p{5cm}}\tabitem Very simple to evaluate and interpret\\ \tabitem Works well for small number of unique answers\end{tabular}                                 & \begin{tabular}[c]{p{5cm}}\tabitem Both minor and major errors are penalized equally\\ \tabitem Can lead to explosion in number of unique answers,\\  \tabitem especially with presence of phrasal or sentence answers\end{tabular} \\ \midrule

\textbf{\begin{tabular}[c]{@{}c@{}}Modified\\ WUPS\end{tabular}}     & \begin{tabular}[c]{p{5cm}}\tabitem More forgiving to simple variations and errors\\ \tabitem Does not require exact match\\ \tabitem Easy to evaluate with simple script\end{tabular} & \begin{tabular}[c]{p{5cm}}\tabitem Generates high scores for answers that are lexically related but have diametrically opposite meaning\\ \tabitem Cannot be used for phrasal or sentence answers\end{tabular}         \\ \midrule

\textbf{\begin{tabular}[c]{@{}c@{}}Consensus \\ Metric\end{tabular}} & \begin{tabular}[c]{p{5cm}}\tabitem Common variances of same answer could be captured\\ \tabitem Easy to evaluate after collecting consensus data\end{tabular}                        & \begin{tabular}[c]{p{5cm}}\tabitem Can allow for some questions having two correct answers\\ \tabitem Expensive to collect ground truth \\ \tabitem Difficulty due to lack of consensus \end{tabular}  \\   \midrule

\textbf{\begin{tabular}[c]{@{}c@{}}Manual \\ Evaluation\end{tabular}} & \begin{tabular}[c]{p{5cm}}\tabitem Variances to same answer is easily captured\\ \tabitem Can work equally well for single word as well as phrase or sentence answers \end{tabular}                        & \begin{tabular}[c]{p{5cm}}\tabitem Can introduce subjective opinion of individual annotators\\ \tabitem Very expensive to setup and slow to evaluate, especially for larger datasets \\ \end{tabular}  \\   

\\ \bottomrule
\end{tabular}
\end{table}

Wu-Palmer Similarity (WUPS)~\cite{wu1994verbs} was proposed as an alternative to accuracy in \cite{malinowski2014multi}. It tries to measure how much a predicted answer differs from the ground truth based on the difference in their semantic meaning. Given a ground truth answer and a predicted answer to a question, WUPS will assign a value between 0 and 1 based on their similarity to each other. It does this by finding the least common subsumer between two semantic senses and assigning scores based on how far back the semantic tree needs to be traversed to find the common subsumer. Using WUPS, semantically similar, but non-identical, words are penalized relatively less. Following our earlier example, `bald eagle' and `eagle' have similarity of 0.96, whereas `bald eagle' and `bird' have similarity of 0.88. However, WUPS tends to assign relatively high scores to even distant concepts, e.g., `raven' and `writing desk' have a WUPS score of 0.4. To remedy this, \cite{malinowski2014multi} proposed to threshold WUPS scores, where a score that is below a threshold will be scaled down by a factor. A threshold of 0.9 and scaling factor of 0.1 was suggested by \cite{malinowski2014multi}. This modified WUPS metric is the standard measure used for evaluating performance on DAQUAR and COCO-QA, in addition to simple accuracy. 

There are two major shortcomings to WUPS that make it difficult to use. First, despite using a thresholded version of WUPS, certain pairs of words are lexically very similar but carry vastly different meaning. This is particularly problematic for questions about object attributes, such as color questions. For example, if the correct answer was `white' and the predicted answer was `black,' the answer would still receive a WUPS score of 0.91, which seems excessively high. Another major problem with WUPS is that it only works with rigid semantic concepts, which are almost always single words. WUPS cannot be used for phrasal or sentence answers that are occasionally found in The VQA Dataset and in much of Visual7W.

An alternative to relying on semantic similarity measures is to have multiple independently collected ground truth answers for each question, which was done for The VQA Dataset~\cite{antol2015vqa} and DAQUAR-consensus~\cite{malinowskiRF15}. For DAQUAR-consensus, an average of five human annotated ground truth answers per question were collected. The dataset's creators proposed two ways to use these answers, which they called average consensus and min consensus. For average consensus, the final score is weighted toward preferring the more popular answer provided by the annotators.  For min consensus, the answer needs to agree with at least one annotator. 

For The VQA Dataset, annotators generated ten answers per question. These are used with a variation of the accuracy metric, which is given by 
\begin{equation}
{Accuracy}_{{VQA}}=\min (\frac{n}{3},1),
\label{eq:vqa-acc}
\end{equation}
where $n$ is the total number of annotators that had the same answer as the algorithm. Using this metric, if the algorithm agrees with three or more annotators then it is awarded a full score for a question. Although this metric helps greatly with the ambiguity problem, substantial problems remain, especially with the COCO-VQA portion of the dataset, which we study further in the next few paragraphs\footnote{Note that our analysis for COCO-VQA was only done on the train and validation portions of the dataset, because the test answers are not publicly available.}. 

Using $Accuracy_{VQA}$, the inter-human agreement on COCO-VQA is only 83.3\%. It is impossible for an algorithm to achieve 100\% accuracy. Inter-human agreement is especially poor for `Why' questions, with over 59\% of these questions having less than three annotators giving exactly the same answer. This makes it impossible to get a full score on these questions. Lack of inter-human agreement can also be seen in simpler, more straightforward questions (see Figure \ref{fig:vqa_disagreement}). In this example, if a system predicts any of the 10 answers, it will be awarded a score of at least 1/3. In several cases, the answers provided by annotators consist complete antonyms (e.g., left and right).

In many other cases, $Accuracy_{VQA}$ leads to multiple correct answers for a question that are in direct opposition to each other. For example, in COCO-VQA more than 13\% of the `yes/no' answers have both `yes' and `no' repeated by more than three annotators. Either answering `yes' or `no' would receive the highest possible score. Even if eight annotators answered `yes,' if two answered `no' then an algorithm would still receive a score of 0.67 for the question. The weight of the majority does not play a role in evaluation.

These problems can result in the scores being inflated. For example, answering `yes' to all yes/no questions should ideally have a score of around 50\% for those questions. However, using $Accuracy_{VQA}$, the score is 71\%. This is partially due to the dataset being biased, with the majority answer for these questions being `yes' 58\% of the time, but a score of 71\% is excessively inflated.

Evaluating the open-ended responses of VQA systems is made simpler when the answers consist of one word answers. This occurs in 87\% of COCO-VQA questions, 100\% of COCO-QA questions, and 90\% of DAQAUR questions. The possibility of multiple correct answers increases greatly when answers need to be multiple words. This occurs frequently in FM-IQA, Visual7W, and Visual Genome, e.g., 27\% of Visual7W answers have three or more words. In this scenario, metrics such as $Accuracy_{VQA}$ are unlikely to help score predicted answers to ground truth answers in open-ended VQA. 

The creators of FM-IQA~\cite{gao2015you} suggested using human judges to assess multi-word answers, but this presents a number of problems. First, using human judges is an extremely demanding process in terms of time, resources, and expenses. It would make it difficult to iteratively improve a system by measuring how changing the algorithm altered performance. Second, human judges need to be given criteria for judging the quality of an answer. The creators of FM-IQA proposed two metrics for human judges. The first is to determine whether the answer was produced by a human or not, regardless of the answer's correctness. This metric alone may be a poor indicator of a VQA system's abilities and could potentially be manipulated. The second metric is to rate an answer on a 3-point scale of totally wrong (0), partially correct (1), and perfectly correct (2). 

An alternative to using judges for handling multi-word answers is to use a multiple-choice paradigm, which is used by part of The VQA Dataset, Visual7W, and Visual Genome. Instead of generating an answer, a system only needs to predict which of the given choices is correct. This greatly simplifies evaluation, but we believe that unless it is used carefully, multiple-choice is ill-suited for VQA because it undermines the effort by allowing a system to peek at the correct answer. We discuss this issue in Section \ref{sec:oeVSmc}.

The best way to evaluate a VQA system is still an open question. Each evaluation method has its own strengths and weaknesses (see Table \ref{table:metrics} for a summary). The method to use depends on how the dataset was constructed, the level of bias within it, and available resources. Considerable work needs to be done to develop better tools for measuring the semantic similarity of answers and for handling multi-word answers.

\section{Algorithms for VQA} \label{sec:previous}

\begin{figure}[t]
\centering 
\includegraphics[width=0.98\textwidth]{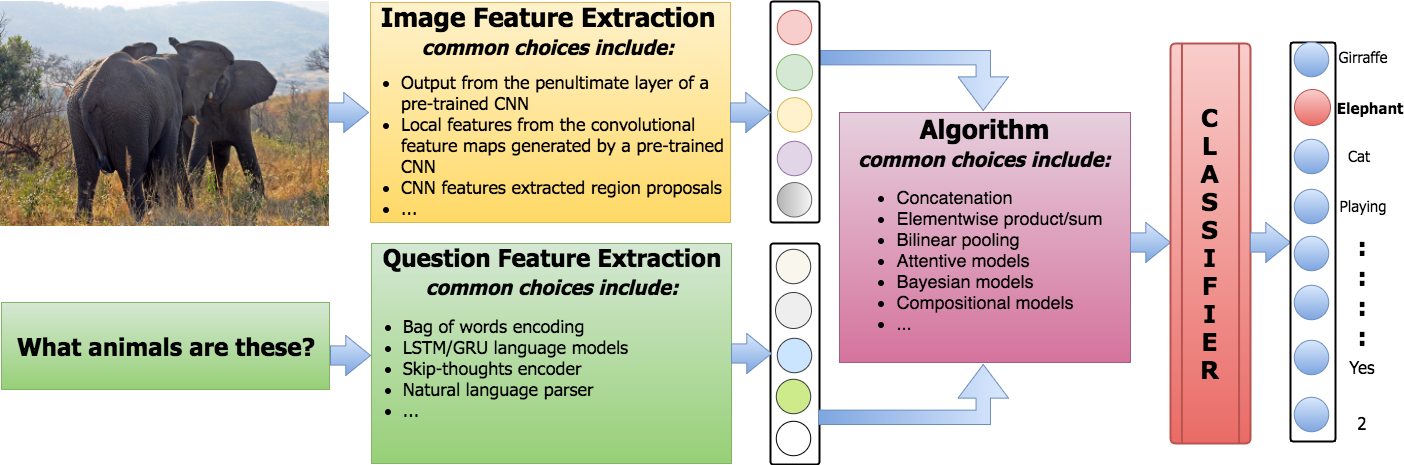}
        \caption{Simplified illustration of the classification based framework for VQA. In this framework, image and question features are extracted, and then they are combined so that a classifier can predict the answer. A variety of feature extraction methods and algorithms for combining these features have been proposed, and some of the more common approaches are listed in their respective blocks in the figure. Full details are presented in Section~\ref{sec:previous}.}
\label{fig:framework}
    \end{figure}

\begin{table}[t]
\centering
\caption{Results across VQA datasets for both open-ended (OE) and multiple-choice (MC) evaluation schemes. Simple models trained only on the image data (IMG-ONLY) and only on the question data (QUES-ONLY) as well as human performance are also shown. IMG-ONLY and QUES-ONLY models are evaluated on the `test-dev' section of COCO-VQA. MCB-ensemble~\cite{FukuiPYRDR16} and AMA~\cite{WuWSHD15} are presented separately as they use additional data for training.}
\small
\label{table:comparison}
\begin{tabular}{@{}lcccccc@{}}
\toprule
\textbf{}      & \multicolumn{2}{c}{\textbf{DAQUAR}}  & \textbf{COCO-QA} &  \multicolumn{2}{c}{\textbf{COCO-VQA}} \\ \midrule
\textbf{}      & \textbf{FULL}    & \textbf{37}      & \textbf{}                & \textbf{OE}    & \textbf{MC}     \\ \midrule
IMG-ONLY~\cite{kafle2016} & \, 6.19 & \, 7.93 & 34.36  & 29.59  & - \\
QUES-ONLY~\cite{kafle2016} & 25.57  & 39.66 & 39.24 & 49.56 & - \\ \midrule
MULTI-WORLD~\cite{malinowski2014multi}    & \, 7.86             & 12.73            & -                               & -              & -               \\
ASK-NEURON~\cite{malinowskiRF15}     & 21.67            & 34.68            & -                                 & -              & -               \\
ENSEMBLE~\cite{ren2015image}       & -                & 36.94            & 57.84                             & -              & -               \\
LSTM Q+I~\cite{antol2015vqa}       & -                & -                & -                                 & 54.06          & 57.17          \\
iBOWIMG~\cite{zhou2015}        & -                 &  -               & -                                    & 55.89          & 61.97           \\
DPPNet~\cite{NohSH15}         & 28.98            & 44.48            & 61.19                             & 57.36          & 62.69           \\
SMem~\cite{xu2015ask}           &                 - & 40.07            &  -                                   & 58.24          &   -              \\
SAN~\cite{Yang2016}            & 29.3~             & 45.5~             & 61.6~                              & 58.9~           & -               \\
NMN~\cite{AndreasRDK15}            & -                & -                & -                                 & 58.7~           & -               \\
D-NMN~\cite{andreas2016learning}          & -                & -                & -                                 & 59.4~           & -               \\
FDA~\cite{ilievski2016focused}            & -                & -                & -                                 & 59.54          & 64.18           \\
HYBRID~\cite{kafle2016}         & 28.96            & 45.17            & 63.18                         & 60.06          & -               \\
DMN+~\cite{xiong2016dynamic}           & -                & -                & -                                 & 60.4~           & -               \\
MRN~\cite{kim2016multimodal}            & -                & -                & -                                 & 61.84          & 66.33           \\
HieCoAtten~\cite{LuYBP16}     & -                & -                & 65.4~                              & 62.1~           & 66.1~            \\
RAU\_ResNet~\cite{noh2016training}    & -                & -                & -                                 & 63.2~           & 67.3~            \\
DAN~\cite{nam2016dual}    & -                & -                & -                                 & 64.2~           & 69.0~            \\
MCB+Att~\cite{FukuiPYRDR16}   & -                & -                & -                                 & 64.2~          & -            \\
MLB~\cite{kim2016hadamard}    & -                & -                & -                                 & 65.07           & 68.89            \\ \midrule
AMA~\cite{WuWSHD15}            &  -                &  -                & 69.73                               & 59.44           &   -              \\
MCB-ensemble~\cite{FukuiPYRDR16}   & -                & -                & -                                 & 66.5~          & 70.1~            \\ 
\midrule
\textbf{HUMAN} & \textbf{50.20}    & \textbf{60.27}   & \textbf{-}            & \textbf{83.30}  & \textbf{91.54}  \\ \bottomrule
\end{tabular}
\end{table}

\begin{table}[t]
\centering
\caption{Overview of different methods that were evaluated on open-ended COCO-VQA and their design choices. Results are report on the  `test-dev' split when `test-standard' results are not available (Denoted by *). }
\label{tbl:choices}
\begin{tabular}{@{}lllcccc@{}}
Method     & \begin{tabular}[c]{@{}l@{}}Accuracy (\%)\\ ($Acc_{VQA}$)\end{tabular} & \begin{tabular}[c]{@{}l@{}}CNN\\ Network\end{tabular} & \begin{tabular}[c]{@{}l@{}}Use of\\Attention\end{tabular} & \begin{tabular}[c]{@{}l@{}}Ext.\\ Data\end{tabular} & \begin{tabular}[c]{@{}l@{}}Compo-\\ sitional\end{tabular} & \\ \midrule
LSTM Q+I~\cite{antol2015vqa}   & 54.1 & VGGNet & - & - & - \\
iBOWIMG~\cite{zhou2015}        & 55.9   & GoogLeNet  & - & - & -           \\
DPPNet~\cite{NohSH15}         & 57.4 & VGGNet & - & - & -            \\
SMem~\cite{xu2015ask}         & 58.2 & GoogLeNet & \checkmark & - & -               \\
SAN~\cite{Yang2016}           & 58.9~ & GoogLeNet &  \checkmark & - & - \\
NMN~\cite{AndreasRDK15}       & 58.7& VGGNet &  \checkmark & - & \checkmark \\
D-NMN~\cite{andreas2016learning} & 59.4~& VGGNet &  \checkmark & - & \checkmark \\
AMA~\cite{WuWSHD15}           & 59.4 & VGGNet & - & \checkmark & - \\
FDA~\cite{ilievski2016focused}  & 59.5 & ResNet &  \checkmark & - \\
HYBRID~\cite{kafle2016}         & 60.1 & ResNet & - & - & -\\
DMN+~\cite{xiong2016dynamic}    & 60.4~ & ResNet &  \checkmark & - & - \\
MRN~\cite{kim2016multimodal}    & 61.8 & ResNet &  \checkmark & - & - \\
HieCoAtten-VGG*~\cite{LuYBP16}     & 60.5~& VGGNet &  \checkmark & - & - \\
HieCoAtten-ResNet~\cite{LuYBP16}     & 62.1~& ResNet &  \checkmark & - & -\\
RAU\_VGG*~\cite{noh2016training}    & 61.3~& VGGNet &  \checkmark & - & - \\
RAU\_ResNet~\cite{noh2016training}    & 63.2~ & ResNet &  \checkmark & -  \\
MCB*~\cite{FukuiPYRDR16}   & 61.2~& ResNet &  - & - & - \\
MCB-ATT*~\cite{FukuiPYRDR16}   & 64.2~& ResNet &  \checkmark & - & - \\
DAN-VGG*~\cite{nam2016dual} & 62.0 & VGGNet & \checkmark & - & - \\
DAN-ResNet~\cite{nam2016dual} & 64.3 & ResNet & \checkmark & - & - \\
MLB~\cite{kim2016hadamard} & 65.1 & ResNet & \checkmark & -& - \\
MLB+VG*~\cite{kim2016hadamard} & 65.8 & ResNet &\checkmark & \checkmark & - \\
MCB-ensemble~\cite{FukuiPYRDR16}   & 66.5~& ResNet &  \checkmark & \checkmark & - \\
\end{tabular}
\end{table}

A large number of VQA algorithms have been proposed in the past three years. All existing methods consist of 1) extracting image features (image featurization), 2) extracting question features (question featurization), and 3) an algorithm that combines these features to produce an answer.  For image features, most algorithms use CNNs that are pre-trained on ImageNet, with common examples being VGGNet~\cite{Simonyan14}, ResNet~\cite{resnet}, and GoogLeNet~\cite{szegedy2015going}. A wider variety of question featurizations have been explored, including bag-of-words (BOW), long short term memory (LSTM) encoders \cite{hochreiter1997long}, gated recurrent units (GRU) \cite{cho2014learning}, and skip-thought vectors \cite{kiros2015skip}. To generate an answer, the most common approach is to treat VQA as a classification problem. In this framework, the image and question features are the input to the classification system and each unique answer is treated as a distinct category.  As illustrated in Figure \ref{fig:framework}, the featurization scheme and the classification system can take widely varied forms. These systems differ significantly in how they integrate the question and image features. Some examples include:
\begin{itemize}[noitemsep] 
\item Combining the image and question features using simple mechanisms, e.g., concatenation, elementwise multiplication, or elementwise addition, and then giving them to a linear classifier or a neural network \cite{kafle2016,zhou2015,antol2015vqa,gao2015you},
\item Combining the image and question features using bilinear pooling or related schemes in a neural network framework \cite{FukuiPYRDR16,kim2016multimodal,saito2016dualnet},
\item Having a classifier that uses the question features to compute spatial attention maps for the visual features or that adaptively scales local features based on their relative importance \cite{Yang2016,ilievski2016focused,xu2015ask,shih2016},
\item  Using Bayesian models that exploit the underlying relationships between question-image-answer feature distributions \cite{kafle2016,malinowski2014multi}, and
\item Using the question to break the VQA task into a series of sub-problems \cite{andreas2016learning,AndreasRDK15}.
\end{itemize}
In later subsections, we describe each of these classification-based approaches in detail.

While the classification framework is used by most open-ended VQA algorithms, this approach can only generate answers that are seen during training, prompting some to explore alternatives.  In \cite{gao2015you} and \cite{malinowskiRF15} an LSTM is used to produce multi-word answer one word at a time. However, the answer produced is still limited to words seen during training. For multiple-choice VQA, \cite{jabri2016revisiting} and \cite{shih2016} proposed treating VQA as a ranking problem, where a system is trained to produced a score for each possible multiple-choice answer, question, and image trio, and then it selects the highest scoring answer choice.

In the following subsections, we group VQA algorithms based on their common themes. Results on DAQUAR, COCO-QA, and COCO-VQA for these methods are given in Table \ref{table:comparison}, in increasing order of performance. In Table \ref{table:comparison}, we report plain accuracy for DAQUAR and COCO-QA, and we report $Accuracy_{VQA}$ for COCO-VQA. Table \ref{tbl:choices} breaks down the results for COCO-VQA based on the techniques used in each  paper.

\subsection{Baseline Models}
Baseline methods help determine the difficulty of a dataset, and establish the minimal level of performance that a more sophisticated algorithms should exceed. For VQA, the simplest baselines are random guessing and guessing the most repeated answers. A widely used baseline classification system is to apply a linear or non-linear, e.g., multi-layer perceptron (MLP), classifier to the image and question features after they have been combined into a single vector~\cite{antol2015vqa,kafle2016,zhou2015}. Common methods to combine the features include concatenation, the elementwise product, or the elementwise sum. Combining these schemes has also been explored and can lead to improved results~\cite{saito2016dualnet}. 

A variety of featurization approaches have been used with baseline classification frameworks. In \cite{zhou2015}, the authors used a bag-of-words to represent the question and CNN features from GoogLeNet for the visual features. They then fed concatenation of these features into a multi-class logistic regression classifier. Their approach worked well, surpassing the previous baseline on COCO-VQA, which used a theoretically more powerful model, an LSTM, to represent the question \cite{antol2015vqa}. Similarly, \cite{kafle2016} used skip-thought vectors \cite{kiros2015skip} for question features and ResNet-152 to extract image features. They found that an MLP model with two hidden layers trained on these off-the-shelf features worked well for all datasets. However, in their work a linear classifier outperformed the MLP model on smaller datasets, likely due to the MLP model overfitting.

Several VQA algorithms have used LSTMs to encode questions. In \cite{antol2015vqa}, an LSTM encoder acting on a one-hot encoding of the sentence was used to represent question features, and GoogLeNet was used for image features. The dimensionality of the CNN features was reduced to match the dimensionality of the LSTM encoding, and then the Hadamard product of these two vectors was used to fuse them together. The fused vector was used as input to an MLP with two hidden layers. In \cite{malinowskiRF15}, an LSTM model was fed an embedding of each word sequentially with CNN features concatenated to it. This continued until the end of the question was reached. The subsequent time-steps were used to generate a list of answers. A related approach was used in \cite{ren2015image}, where an LSTM was fed CNN features during the first and last time-steps, with word features in between. The image features acted as the first and last words in the sentence. The LSTM network was followed by a softmax classifier to predict the answer. A similar approach was used in \cite{gao2015you}, but the CNN image features were only fed into the LSTM at the end of the question and instead of a classifier, another LSTM was used to generate the answer one word at a time.
\subsection{Bayesian and Question-Aware Models}
VQA requires drawing inferences and modeling relationships between the question and the image. Once the questions and images are featurized, modeling co-occurrence statistics of the question and image features can be helpful for drawing inferences about the correct answers. Two major Bayesian VQA frameworks have explored modeling these relationships. In \cite{malinowski2014multi}, the first Bayesian framework for VQA was proposed. The authors used semantic segmentation to identify the objects in an image and their positions. Then, a Bayesian algorithm was trained to model the spatial relationships of the objects, which was used to compute each answer's probability. This was the earliest known algorithm for VQA, but its efficacy is surpassed by simple baseline models. This is partially due to it being dependent on the results of the semantic segmentation, which was imperfect.

A very different Bayesian model was proposed in \cite{kafle2016}. The model exploited the fact that the type of answer can be predicted using solely the question. For example, `What color is the flower?' would be assigned as a color question by the model, essentially turning the open-ended problem into a multiple-choice one. To do this, the model used a variant of quadratic discriminant analysis, which modeled the probability of image features given the question features and the answer type. ResNet-152 was used for the image features, and skip-thought vectors were used to represent the question.

\subsection{Attention Based Models}
\begin{figure*}[t]
	\centering \par\medskip
		\includegraphics[width=\textwidth]{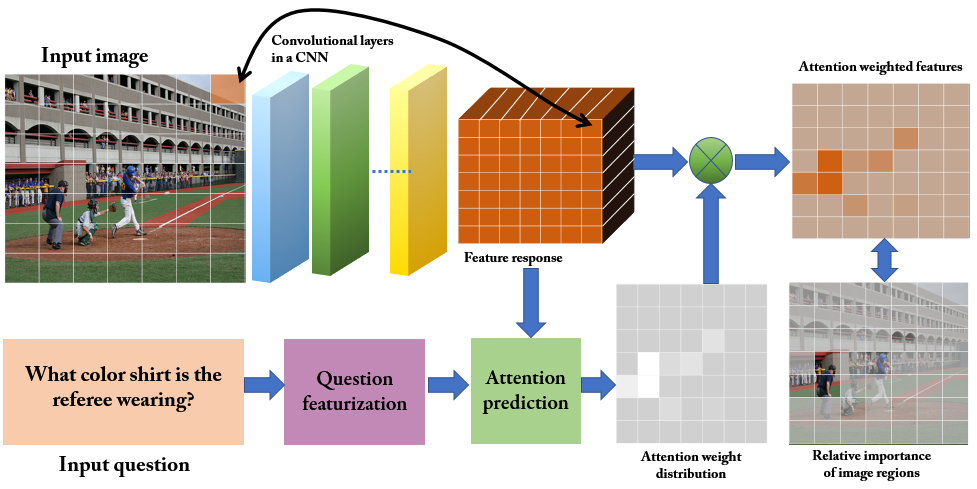}        \hfill    
    \caption{This figure illustrates a common way to incorporate attention into a VQA system. A convolutional layer in a CNN outputs a $K \times K \times N$ tensor of feature responses, corresponding to $N$ feature maps. One way to apply attention to this representation is by suppressing or enhancing the features at different spatial locations. Using the question features with these local image features, a weighting factor for each grid location can be computed that determines the spatial location's relevance to the question, which can then be used to compute attention-weighted image features.}
    \label{fig:attention} 
\end{figure*}

Using global features alone may obscure task-relevant regions of the input space. Attentive models attempt to overcome this limitation. These models learn to `attend' to the most relevant regions of the input space. Attention models have shown great successes in other vision and NLP tasks, such as object recognition~\cite{ba2014multiple}, captioning~\cite{xu2015show} and machine translation~\cite{luong2015effective,bahdanau2014neural}.

In VQA, numerous models have used spatial attention to create region-specific CNN features, rather than using global features from the entire image. Fewer models have also explored incorporating attention into the text representation. The basic idea behind all these models is that certain visual regions in an image and certain words in a question are more informative than others for answering a given question. For example, for a system answering `What color is the umbrella?' the image region containing the umbrella is more informative than other image regions. Similarly, `color' and `umbrella' are the textual inputs that need to be addressed more directly than the others. Global image features, e.g., the last hidden layer of a CNN, and global text features, e.g., bag-of-words, skip-thoughts etc. may not be granular enough to address region specific questions. 

Before using spatially attentive mechanisms, an algorithm must represent the visual features at all spatial regions, instead of solely at the global level. Then, local features from relevant regions can be given higher prominence based on the question asked. There are two common ways to achieve local feature encoding. As shown in Figure \ref{fig:attention}, one way to do this is to impose a uniform grid over all image locations, with the local image features present at each grid location. This is often done by operating on the last CNN layer prior to the final spatial pooling that flattens the features. The relevance of each grid location is then determined by the question. An alternative way to implement spatial attention is to generate region proposals (bounding boxes) for an image, encode each of these boxes using a CNN, and then determine the relevance of each box's features using the question. While multiple papers have focused on using spatial visual attention for VQA~\cite{shih2016,Yang2016,xiong2016dynamic,xu2015ask,LuYBP16,ilievski2016focused,FukuiPYRDR16,noh2016training}, there are significant differences among these methods.

The Focus Regions for VQA~\cite{shih2016} and Focused Dynamic Attention (FDA) models~\cite{ilievski2016focused} both used Edge Boxes~\cite{zitnick2014edge} to generate bounding box region proposals for images. In \cite{shih2016}, a CNN was used to extract features from each of these boxes. The input to their VQA system consisted of these CNN features, question features, and one of the multiple choice answers. Their system was trained to produced a score for each multiple-choice answer, and the highest scoring answer was selected. The score is calculated using a weighted average of scores from each of the regions where the weights are simply learned by passing the dot product of regional CNN feature and question embedding to a fully connected layer.

In FDA~\cite{ilievski2016focused}, the authors proposed to only use the region proposals that have the objects mentioned in the question. Their VQA algorithm requires as input a list of bounding boxes with their corresponding object label. During training, the object labels and bounding boxes are obtained from COCO annotations. During test, the labels are obtained by classifying each bounding box using ResNet~\cite{resnet}. Subsequently, word2vec~\cite{mikolov2013distributed} was used to compute the similarity between words in the question and the object labels assigned to each of the bounding boxes. Any box with a score greater than 0.5 is successively fed into an LSTM network. At the last time-step, global CNN features from the entire image are also fed into the network, giving it access to both global and local features. A separate LSTM was also used as the question representation. The output from these two LSTMs are then fed into a fully connected layer that is fed to a softmax classifier to produce the answer predictions.

In contrast to using region proposals, the Stacked Attention Network (SAN)~\cite{Yang2016} and the Dynamic Memory Network (DMN)~\cite{xiong2016dynamic} models both used visual features from the spatial grid of a CNN's feature maps (see Figure \ref{fig:attention}). Both \cite{Yang2016} and \cite{xiong2016dynamic} used the last convolutional layer from VGG-19 with $448 \times 448$ images to produce a $14 \times 14$ filter response map with 512 dimensional features at each grid location.

In SAN~\cite{Yang2016}, an attention layer is specified by a single layer of weights that uses the question and the CNN feature map with a softmax activation function to compute the attention distribution across image locations. This distribution is then applied to the CNN feature map to pool across spatial feature locations using a weighted sum, which generates a global image representation that emphasizes certain spatial regions more than others. This feature vector is then combined with a vector of question features to create a representation that can be used with a softmax layer to predict the answer. They generalized this approach to handle multiple (stacked) attention layers, enabling the system to model complex relationships among multiple objects in an image.
 
A similar attentive mechanism was used in the Spatial Memory Network~\cite{xu2015ask} model, where spatial attention is produced by estimating the correlation of image patches with individual words in the question. This word-guided attention is used to predict an attention distribution, which is then used to compute the weighted sum of the visual features embedding across image regions. Two different models were then explored. In the one-hop model, the features encoding the entire question are combined with the weighted visual features to predict the answer. In the two-hop model, the combination of the visual and question features is looped back into the attentive mechanism for refining the attention distribution.

Another approach that incorporated spatial attention using CNN feature maps is presented in \cite{xiong2016dynamic}. To do this, they used a modified Dynamic Memory Network (DMN)~\cite{kumar2015ask}. A DMN consists of an input module, an episodic memory module, and an answering module. DMNs have been used for text based QA, where each word in a sentence is fed into a recurrent neural network and the output of the network is used to extract `facts.' Then, the episodic memory module makes multiple passes over a subset of these facts. With each pass, the internal memory representation of the network is updated. An answering module uses the final state of the memory representation and the input question to predict an answer. To use a DMN for VQA, they used visual facts in addition to text. To generate visual facts, the CNN features at each spatial grid location are treated as words in a sentence that are sequentially fed into a recurrent neural network. The episodic memory module then makes passes through both text and visual facts to update its memory. The answering module remains unchanged.

The Hierarchical Co-Attention model~\cite{LuYBP16} applies attention to both the image and question to jointly reason about the two different streams of information. The model's approach to visual attention is similar to the method used in Spatial Memory Network~\cite{xu2015ask}. In addition to visual attention, this method uses a hierarchical encoding of the question, in which the encoding occurs at the word level (using a one-hot encoding), at the phrase level (using bi- or tri-gram window size), and at the question level (using the final time-step of an LSTM network). Using this hierarchical question representation, the authors proposed to use two different attentive mechanisms. The parallel co-attention approach simultaneously attended to both the question and image. The alternative co-attention approach alternated between attending to the question or the image. This approach allowed the relevance of words in the question and the relevance of specific image regions to be determined by each other. The answer prediction is made by recursively combining the co-attended features from all three levels of the question hierarchy.

Using joint attention for image and question features was also explored in \cite{nam2016dual}. The main idea is to allow image and question attention to guide each other, directing attention to relevant words and visual regions simultaneously. To achieve this, visual and question input are jointly represented by a memory vector that is used to simultaneously predict attention for both question and image features. The attentive mechanism computes updated image and question representations, which are then used to recursively update the memory vector. This recursive memory update mechanism can be repeated $K$ times to refine the attention in multiple steps. The authors' found that a value of $K=2$ worked best for COCO-VQA.

\subsection{Bilinear Pooling Methods}
VQA relies on jointly analyzing the image and the question. Early models did this by combining their respective features using simple methods, e.g., concatenation or using an element-wise product between the question and image features, but more complex interactions would be possible with an outer-product between these two streams of information. Similar ideas were shown to work well for improving fine-grained image recognition ~\cite{lin2015bilinear}. Below, we describe the two most prominent VQA methods that have used bilinear pooling~\cite{FukuiPYRDR16,kim2016hadamard}.

In \cite{FukuiPYRDR16}, Multimodal Compact Bilinear (MCB) pooling was proposed as a novel method for combining image and text features in VQA. This idea is to approximate the outer-product between the image and text features, allowing a deeper interaction between the two modalities, compared to other mechanisms, e.g., concatenation or element-wise multiplication. Rather than doing the outer-product explicitly, which would be very high dimensional, MCB does the outer-product in a lower dimensional space. This is then used to predict which spatial features are relevant to the question. In a variation of this model, a soft-attention mechanism, similar to the method in \cite{Yang2016}, was also used, with the only major change being the use of MCB for combining text and question features instead of element-wise multiplication in \cite{Yang2016}. This combination yielded very good results on COCO-VQA, and it was the winner of the 2016 VQA Challenge workshop.

In \cite{kim2016hadamard}, the authors' argued that MCB is too computationally expensive, despite using an approximate outer-product. Instead, they proposed to use a multi-modal low-rank bilinear pooling (MLB) scheme that uses the Hadamard product and a linear mapping to achieve approximate bilinear pooling. When used with a spatial visual attention mechanism, MLB rivaled MCB at VQA, but with reduced computational complexity and using a neural network with fewer parameters.

\subsection{Compositional VQA Models}

In VQA, questions often require multiple steps of reasoning to answer properly. For example, questions like `What is to the left of the horse?' can involve first finding the horse, and then naming the object to the left of it. Two compositional frameworks have been proposed for VQA that attempt to tackle solving VQA in a series of sub-steps~\cite{AndreasRDK15,andreas2016learning,noh2016training}. The Neural Module Network (NMN)~\cite{AndreasRDK15,andreas2016learning} framework uses external question parsers to find the sub-task in the question whereas Recurrent Answering Units (RAU)~\cite{noh2016training} is trained end-to-end and sub-tasks can be implicitly learned.

NMN is an especially interesting approach to VQA~\cite{AndreasRDK15,andreas2016learning}. The NMN framework treats VQA as a sequence of sub-tasks that are carried out by separate neural sub-networks. Each of the sub-network performs a single well-defined task, e.g., the \texttt{find[X]} module produces a heat map for the presence of certain object. Other modules include \texttt{describe}, \texttt{measure}, and \texttt{transform}. These modules then must be assembled into a meaningful layout. Two methods have been explored for inferring the required layout. In \cite{AndreasRDK15}, a natural language parser is used on the input question to both find the sub-tasks in the question and to infer the required layout of the sub-tasks that when executed in sequence would produce an answer to the given question \cite{AndreasRDK15}. For example, answering `What color is the tie?' would involve executing the \texttt{find[tie]} module followed by the \texttt{describe[color]} module, which generates the answer. In \cite{andreas2016learning}, the same group explored using algorithms to dynamically select the best layout for the given question from a set of automatically generated layout candidates.

The RAU model~\cite{noh2016training} can implicitly perform compositional reasoning without depending on an external language parser. In their model, they used  multiple self-contained answering units that can solve VQA sub-tasks. These answering units are arranged in recurrent manner. Each answering unit on the chain is equipped with an attentive mechanism derived from \cite{Yang2016} and a classifier. The authors' claimed that the inclusion of multiple recurrent answering units allows inferring the answer from a series of sub-tasks solved by each answering unit. However, they did not perform visualization or ablation studies to show how the answer might get refined in each time-step. This makes it difficult to assess whether progressive refinement and reasoning is occurring or not, especially considering that the complete image and question information is available to all answering units at every time step and that only the output from the first answering unit is used during the test stage.
\subsection{Other Noteworthy Models}
Answering questions about images can often require information beyond what can be directly inferred by analyzing the image. Having knowledge about the uses and typical context for the objects present in an image can be helpful for VQA. For example, a VQA system that has access to a knowledge bank could use it to answer questions about particular animals, such as their habitats, colors, sizes, and feeding habits.  This idea was explored in \cite{WuWSHD15}, and they demonstrated that the knowledge bank improved performance. The external knowledge bases were tailored to general information obtained from DBpedia~\cite{lehmann2015dbpedia}, and it is possible that using a source tailored to VQA could yield greater improvement.

In \cite{NohSH15}, the authors' incorporated a Dynamic Parameter Prediction layer into the fully connected layers of a CNN. The parameters of this layer are predicted from the question by using a recurrent neural network. This allows the visual features that the model uses to be specific to the question before the final classification step. This approach can be seen as a kind of implicit attentive mechanism in that it modifies the visual input based on the question.

In \cite{kim2016multimodal}, Multimodal Residual Networks (MRN) were proposed for VQA, which were motivated by the success of the ResNet architecture in image classification. Their system is a modification of ResNet~\cite{resnet} to use both visual and question features in the residual mapping. The visual and question embedding are allowed to have their own residual blocks with skip connections. However, after each residual block the visual data is inter-weaved with the question embedding. The authors explored several alternate arrangement for constructing the residual architecture with multi-modal input and chose the above network based on performance.  
\subsection{What methods and techniques work better?}
Although many methods have been proposed for VQA, it is difficult to determine what general techniques are superior.  Table \ref{tbl:choices} provides a breakdown of the different algorithms evaluated on COCO-VQA based on the techniques and design choices that they utilize. Table \ref{tbl:choices} also includes ablation models from respective algorithms, whenever possible. The ablation models help us to identify the individual contributions of the design choices made by the authors. The first observation we can make is that ResNet produces superior performance over VGGNet or GoogLeNet across multiple algorithms. This is evident from the models that use identical setup and only change the image representation.  In \cite{noh2016training}, an increase of 2\% was observed by using ResNet-101 instead of the VGG-16 CNN for image features. In \cite{LuYBP16}, they found an increase of 1.3\% when making the same change in their model. Similarly, changing VGG-19 to ResNet-152 increased performance by 
2.3\% in \cite{nam2016dual}. In all cases, rest of the architecture was kept unchanged. 

In general, we believe that spatial attention can be used to increase performance for a model. This is shown by experiments in \cite{FukuiPYRDR16} and \cite{LuYBP16}, where the models were evaluated  with and without attention, and the attentive version performed better in both cases. However, attention alone does not appear to be sufficient. We further discuss this in Section \ref{subsec:attention}.

Bayesian and compositional architectures do not significantly improve over comparable models, despite being interesting approaches. In \cite{kafle2016}, the model performed competitively only after it was combined with an MLP model. It is unclear whether the increase was due to model averaging or the proposed Bayesian method. Similarly, the NMN models in \cite{AndreasRDK15} and \cite{andreas2016learning} do not outperform comparable non-compositional models, e.g., \cite{Yang2016}. It is possible that both of these methods perform well on specific VQA sub-tasks, e.g., NMN was shown to be specially helpful for positional reasoning questions in the SHAPES dataset. However, since major datasets do not provide a detailed breakdown of question types, it is not possible to quantify how systems perform on specific question types. Moreover, any improvements on rare question types will have negligible impact on the overall performance score, making it difficult to properly evaluate the benefits of these methods. We further discuss these issues in Section \ref{subsec:bias_eval}.
\section{Discussion}
\label{sec:discussion}

\begin{figure}[t]
	\centering \par\medskip
		\includegraphics[width=\textwidth]{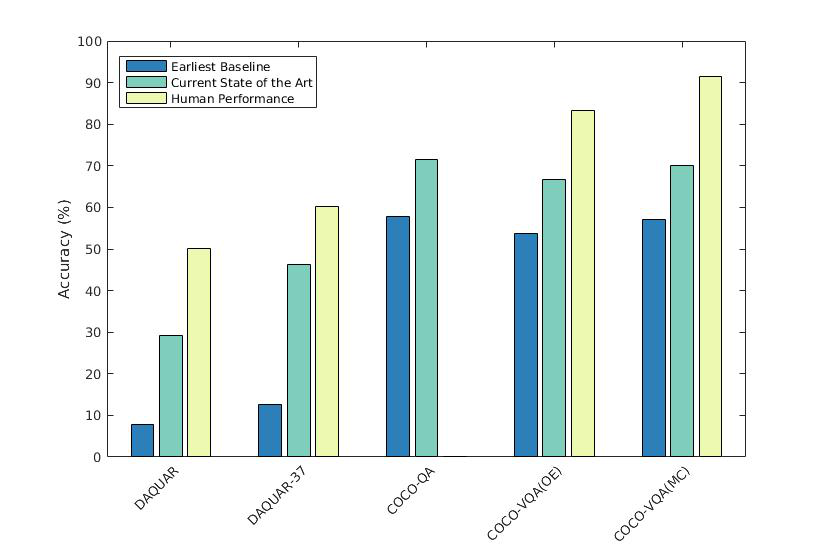}        \hfill
        \caption{Current state-of-the-art results across datasets compared to the earliest baseline and human performance. The earliest baseline refers to the numbers reported by the creators of the datasets and the current state-of-the-art models are chosen from the highest performing methods in Table \ref{table:comparison}. DAQUAR, DAQUAR-37 and COCO-QA report plain accuracy and COCO-VQA reports $Accuracy_{VQA}$.}
        \label{fig:increment}     
\end{figure}

As shown in Figure \ref{fig:increment}, there has been rapid improvement in the performance of VQA algorithms, but there is still a significant gap between the best methods and humans. It remains unclear whether the improvements in performance come from the mechanisms incorporated into later systems, e.g., attention, or if it is due to other factors. Moreover, it can be difficult to decouple the contributions of text and image data in isolation. There are also numerous challenges to comparing algorithms due to the variations in how they are evaluated. In this section, we discuss each of these issues.

\subsection{Vision vs. Language in VQA}
\label{subsec:bias}

VQA consists of two distinct data streams that need to be correctly used to ensure robust performance: images and questions. But, do current systems adequately use both vision and language? Ablation studies~\cite{kafle2016,antol2015vqa} have routinely shown that question only models perform drastically better than image only models, especially on open-ended COCO-VQA. On COCO-QA, simple image-blind models that use only the question can achieve 50\% accuracy with the gain from using the image being comparatively modest~\cite{kafle2016}. In \cite{kafle2016}, it was also shown that for DAQUAR-37, using a better language embedding with an image-blind model produced results superior to earlier works employing both images and questions. This is primarily due to two factors. First, the question severely constrains the kinds of answers expected in many cases, essentially turning an open-ended question into a multiple-choice one, e.g., questions about the color of an object will have a color as an answer. Second, the datasets tend to have strong bias. These two factors make language a much stronger prior than the image features alone. 

    \begin{figure}[t]
    \captionsetup[subfigure]{labelformat=empty}
	\centering
		\begin{subfigure}{0.48\textwidth}
		\includegraphics[width=\textwidth]{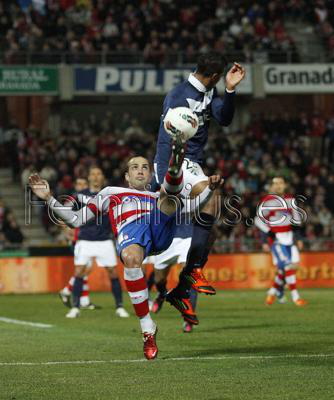}
		 \caption{\textbf{Q:} What are they doing? \textbf{A:} Playing baseball \\ \textbf{Q:} What are they playing? \textbf{A:} Soccer}
		\end{subfigure}
        \hfill
        \begin{subfigure}{0.48\textwidth}
		\includegraphics[width=0.8\textwidth]{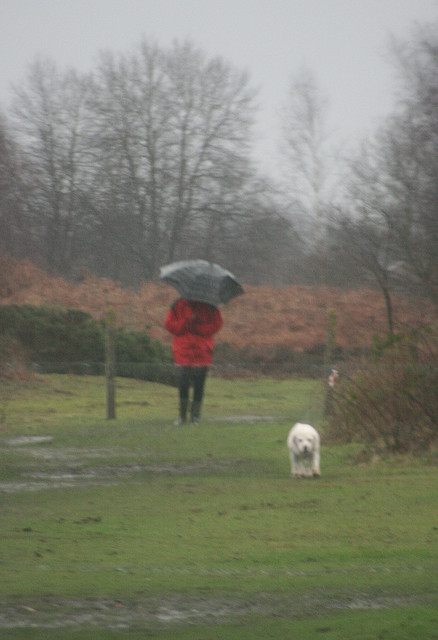}
                \caption{\textbf{Q:} Is the weather rainy in the picture? \textbf{A:} Yes \\ \textbf{Q:} Is it rainy in the picture? \textbf{A:} No }
		\end{subfigure}
        
        \caption{Slight variations in the way a question is phased causes current VQA algorithms to produce different answers. The left example uses the system in \cite{zhou2015} and the right example uses the system from \cite{kafle2016}.}        
         \label{fig:samequestion}
    \end{figure}

The predictive power of language over images have been corroborated by ablation studies. In \cite{AgrawalBP16}, the authors studied a model that had been trained using both image and question features. They then studied how the predictions of the model differed when it was given only the image or only the question, compared to when it was given both. They found that the image-only model's predictions differed from the combined model 40\% more often than the question only model. They also showed that the way the question is phrased strongly biases the answer. When training a neural network, these regularities will be incorporated into the model. While this produces increased performance on the dataset, it is potentially detrimental to creating a general VQA system.

In \cite{Zhang_2016_CVPR}, bias in VQA was studied using synthetic cartoon images. They created a dataset with solely binary questions, in which the same question could be asked about two images that were mostly identical, except for a minor change that caused the correct answer to be different. They found that a model trained on an unbalanced version of this dataset performed 11\% worse (absolute difference) on a balanced test dataset compared to a model trained on a balanced version of the dataset.

\begin{figure}[t]
    \captionsetup[subfigure]{labelformat=empty}
	\begin{subfigure}{0.5\textwidth}
		\includegraphics[width=0.95\textwidth]{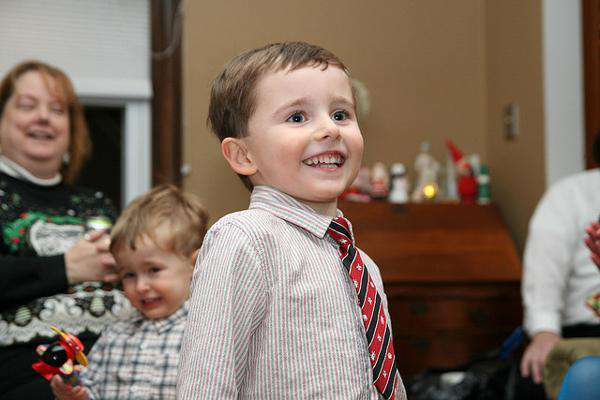}
        \caption{\textbf{no} (11.07 w/ 2.57 [image] + 8.50 [word])\\ \textbf{yes} (10.94 w/ 2.71 [image] + 8.23 [word])}
		\end{subfigure}
        \hfill
        \begin{subfigure}{0.45\textwidth}
		\includegraphics[width=0.95\textwidth]{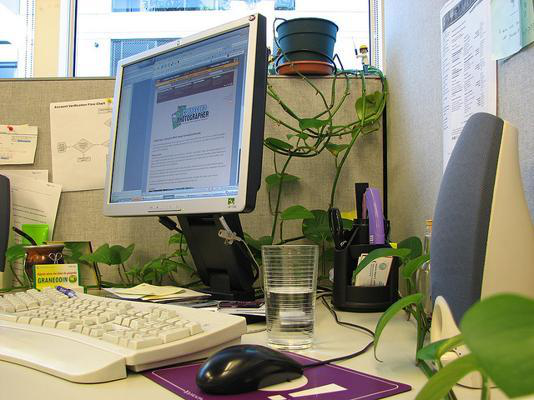}
                \caption{\textbf{yes} (12.45 w/ 4.22 [image] + 8.23 [word])\\ \textbf{no} (12.05 w/ 3.55 [image] + 8.50 [word])}
		\end{subfigure}
        \begin{subfigure}{0.5\textwidth}
		\includegraphics[width=0.95\textwidth]{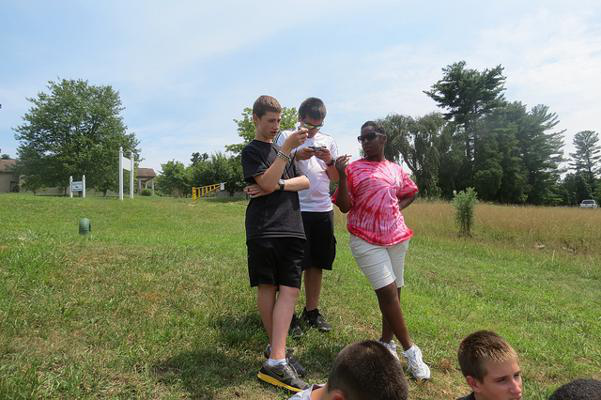}
        \caption{\textbf{no} (12.04 w/ 3.54 [image] + 8.50 [word])\\ \textbf{yes} (11.96 w/ 3.72 [image] + 8.23 [word])}
		\end{subfigure}
        \hfill
        \begin{subfigure}{0.45\textwidth}
		\includegraphics[width=0.95\textwidth]{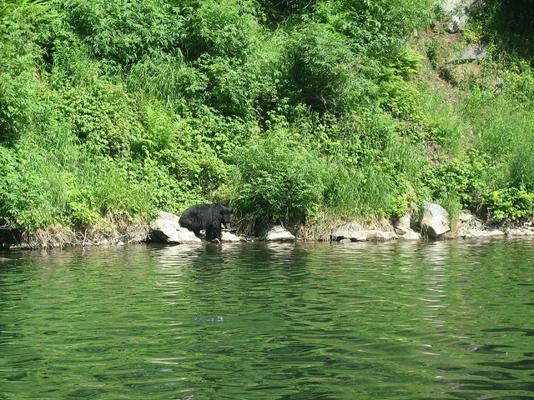}
                \caption{\textbf{yes} (12.30 w/ 4.07 [image] + 8.23 [word])\\ \textbf{no} (12.14 w/ 3.64 [image] + 8.50 [word])}
		\end{subfigure}
   
    \caption{Using the system in \cite{zhou2015}, the answer score for the question `Are there any people in the picture?' is roughly the same for `yes' (8.23) and `no' (8.50). Answering the question correctly requires examining the image, but the model fails to appropriately use the image information.}
    \label{fig:bias}
\end{figure}

We conducted two experiments to assess the effect of language bias in VQA. First, we used the model\footnote{An online demo is available here: \url{http://visualqa.csail.mit.edu/}} from  \cite{zhou2015}. This model was trained on COCO-VQA, and it allows the contribution of the question and image features to be assessed independently by splitting the weights of the softmax output layer into image and question components. We asked simple binary questions with a relatively equal prior for both choices so that the image must be analyzed to answer the question. Examples are shown in Figure \ref{fig:bias}. We can see that the system performs poorly, especially when considering that the baseline accuracy for yes/no questions for COCO-VQA is about 80\%. Next, we studied how language bias affected the more complex MCB-ensemble model \cite{FukuiPYRDR16} that was trained on COCO-VQA. This model was the winner of the 2016 VQA Challenge workshop. To do this, we created a small dataset consisting only of yes/no questions. To create this dataset, we used annotations from the validation split of the COCO dataset to determine whether an image contained a person, and then asked an equal number of `yes' and `no' questions about whether there are any people present. We used the questions `Are there any people in the photo?', `Is there a person in the picture?', and `Is there a person in the photo?' For each variation, there were 38,514 yes/no questions (115,542 total). The accuracy of MCB-ensemble on this dataset was worse than chance (47\%), which starkly contrasts with its results on COCO-VQA (i.e., 83\% on COCO-VQA yes/no questions). This is likely due to severe bias in the training dataset, and not due to an inability for MCB to learn the task.

As shown in Figure \ref{fig:samequestion}, VQA systems are sensitive to the way a question is phrased. We observed similar results when using the system in \cite{antol2015vqa}. To quantify this issue, we created another toy dataset from the validation split of the COCO dataset and used it to evaluate the MCB-ensemble model that was trained on COCO-VQA. In this toy dataset, the task is to identify which sport was being played. We asked three variations of the same question: 1) `What are they doing?', 2) `What are they playing?', and 3) `What sport are they playing?' Each variation contains 5,237 questions about seven common sports (15,711 questions total). MCB-ensemble achieved 33.6\% for variation 1, 78\% for variation 2, and 86.4\% for variation 3.  The dramatic increase in performance from variation 1 to 2 is caused by the inclusion of keyword `playing' instead of the generic verb `doing.' The increment from variation 2 to 3 is caused by explicitly including the keyword `sport' in the question. This suggests that VQA systems are over-dependent on language `clues' that annotators often include. Taken together, these experiments show that language bias is an issue that critically affects the performance of current VQA systems.

In conclusion, current VQA systems are more dependent on the question than the image content. Language bias in datasets critically affects the performance of the current VQA systems, which limits their deployment. New VQA datasets must endeavor to compensate for this issue, by either having questions that force analysis of image content and/or by making datasets less biased.

\subsection{How useful is attention for VQA?}
\label{subsec:attention}
It is difficult to determine how much attention helps VQA algorithms. In ablation studies, when attentive mechanisms are removed from models it impairs their performance~\cite{FukuiPYRDR16,LuYBP16}. Currently, the best model for COCO-VQA does employ spatial visual attention~\cite{FukuiPYRDR16}, but simple models that do not use attention have been shown to exceed earlier models that used complex attentive mechanisms. In \cite{saito2016dualnet}, for example, an attention-free model that used multiple global image feature representations (VGG-19, ResNet-101, and ResNet-152), instead of a single CNN, performed very well compared some attentive models. They combined image and question features using both element-wise multiplication and addition, instead of solely concatenating them. Combined with ensembling, this yielded results significantly higher than the complex attention-based models used in \cite{Yang2016} and \cite{xiong2016dynamic}. Similar results have been obtained by other systems that do not employ spatial attention, e.g,~\cite{kafle2016,jabri2016revisiting,kim2016multimodal}. Attention alone does not ensure good VQA performance, but incorporating attention into a VQA model appears to improve performance over the same model when attention is not used.

In \cite{das2016human}, the authors showed that methods commonly used to incorporate spatial attention to specific image features do not cause models to attend to the same regions as humans tasked with VQA. They made this observation using both the attentive mechanisms used in \cite{Yang2016} and \cite{LuYBP16}. This may be because the regions the model learns to attend to are discriminative due to biases in the dataset and not due to where the algorithm should attend. For example, when asked a question about whether drapes are in an image, the algorithm may instead look at the bottom of the image for a bed rather than windows because questions about drapes tend to be found in bedrooms. This is an indication that attentive mechanisms may not be correctly deployed due to biases.

\subsection{Bias Impairs Method Evaluation}
\label{subsec:bias_eval}

Dataset bias significantly impairs the ability to evaluate VQA algorithms. Questions that require the use of the image content are often relatively easy to answer. Many are about the presence of objects or scene attributes. These questions tend to be handled well by CNNs and also have strong language biases. Harder questions, such as those beginning with `Why' are comparatively rare. This has serious implications for evaluating performance. For COCO-VQA (train and validation partitions), a system that improves accuracy on questions beginning with `Is' and 'Are' by 15\% will increase overall accuracy by 5\%. However, the same increase in both `Why' and `Where' questions will only increase accuracy by 0.6\%. In fact, even if all `Why' and `Where' questions are answered correctly, the overall increase in accuracy will only be 4.1\%. On the other hand, answering `yes' to all questions beginning with `Is there' will yield an accuracy of 85.2\% on those questions. These problems could be overcome if each \textit{type} of question was evaluated in isolation, and then the mean accuracy across question types was used instead of overall accuracy for benchmarking the algorithms. This approach is similar to the mean per-class accuracy metric used for evaluating object classification algorithms, which was adopted due to bias in the amount of test data available for different object categories.

\subsection{Are Binary Questions Sufficient?}
\label{sec:binary-questions}

Using binary (yes/no or true/false) questions to evaluate algorithms has attracted significant discussion in the VQA community. The main argument against using binary questions is the lack of complex questions and the relative ease in answering the questions that are typically generated by human annotators. Visual Genome and Visual7W exclude binary questions altogether. The authors argued that this choice would encourage more complex questions from the annotators.

On the other hand, binary questions are easy to evaluate and these questions can, in theory, encompass an enormous variety of tasks. The SHAPES dataset~\cite{AndreasRDK15} uses binary questions exclusively but contains complex questions involving spatial reasoning, counting, and drawing inferences (see Figure \ref{fig:shapes}). Using cartoon images, \cite{Zhang_2016_CVPR} also showed that these questions can be especially difficult for VQA algorithms when the dataset is balanced. However, there are challenges to creating balanced binary questions for real world imagery. In COCO-VQA, `yes' is a much more common answer than `no,' simply because people tend to ask questions biased toward `yes' as an answer.

As long as bias is controlled, yes/no questions can play an important role in future VQA benchmarks, but a VQA system should be capable of more than solely binary questions so that its abilities can be fully assessed. All real-world applications for VQA, such as enabling the blind to ask questions about visual content, require the output of the VQA system to be open-ended. A system that can solely handle binary questions will have limited real-world utility.

\subsection{Open Ended vs. Multiple Choice} \label{sec:oeVSmc}
Because it is challenging to evaluate open-ended multi-word answers, multiple-choice has been proposed as a way to evaluate VQA algorithms. As long as the alternatives are sufficiently difficult, a system could be evaluated in this manner but then be deployed to answer open-ended questions. For these reasons, multiple choice is used to evaluate Visual7W, Visual Genome, and a variant of The VQA Dataset. In this framework, an algorithm has access to a number of possible answers (e.g., 18 for COCO-VQA), along with the question and image. It must then select among possible choices. 

A major problem with multiple-choice evaluation is that the problem can be reduced to determining which of the answers is correct instead of actually answering the question. For example, in \cite{jabri2016revisiting}, they formulated VQA as an answer scoring task, where the system was trained to produce a score based on the image, question, and potential answers. The answers themselves were fed into the system as features. It achieved state-of-the-art results on Visual7W and rivals the best methods on COCO-VQA, with their method performing better than many complex systems that use attention. To a large extent, we believe their system performed well because it learned to better exploit biases in the answers instead of reasoning about images. On Visual7W, they showed that a variant of their system that used solely the answers and was both image- and question-blind rivaled baselines using the question and image.

We argue that any VQA system should be able to operate without being given answers as inputs. Multiple-choice can be an important ingredient for evaluating multi-word answers, but it alone is not sufficient. When multiple-choice is used, the choices must be selected carefully to ensure that a question is hard and not deducible from the provided answers alone. A system that is solely capable of operating with answers provided is not really solving VQA, because these are not available when a system is deployed. 

\section{Recommendations for Future VQA Datasets}
\label{sec:future}
\begin{figure}
	\centering
\includegraphics[width=0.75\textwidth]{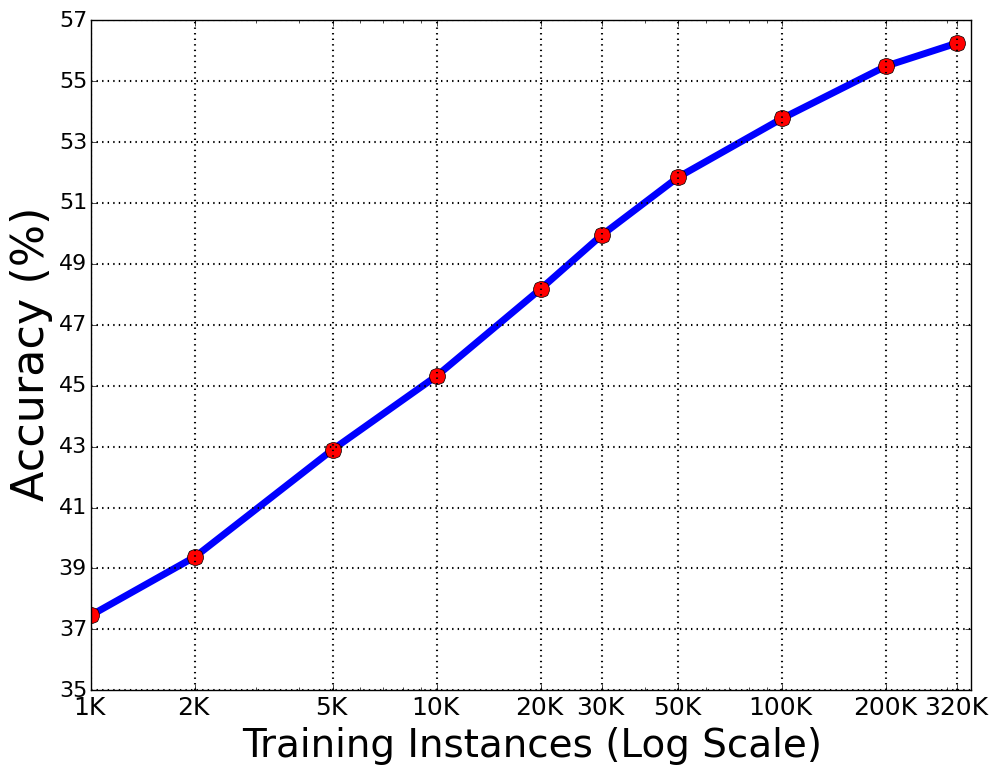} 
    \caption{The graph showing test accuracy as a function of available training data on the COCO-VQA dataset.}
    \label{fig:limited_training}
\end{figure}

Existing VQA benchmarks are not sufficient to evaluate whether an algorithm has `solved' VQA. In this section, we discuss future developments in VQA datasets that will make them better benchmarks for the problem.

Future datasets need to be larger. While VQA datasets have been growing in size and diversity, algorithms do not have enough data for training and evaluation. We did a small experiment where we trained a simple MLP baseline model for VQA using ResNet-152 image features and skip-thought features for the questions, and we assessed performance as a function of the amount of training data available on COCO-VQA. The results are shown in Figure~\ref{fig:limited_training}, where it is clear that the curve has not started to approach an asymptote. This suggests that even on datasets that are biased, increasing the size of the dataset could significantly improve accuracy. However, this does not mean that increasing the size of the dataset is sufficient to turn it into a good benchmark, because humans tend to create questions with strong biases. 

Future datasets need to be less biased. We have repeatedly discussed the problem of bias in existing VQA datasets in this paper, and pointed out the kinds of problems these biases cause for truly evaluating a VQA algorithm. For real-world open-ended VQA, this will be difficult to achieve without carefully instructing the humans that generate the questions. Bias has long been a problem in images used for computer vision datasets (for a review see \cite{torralba2011}), and for VQA this problem is compounded by bias in the questions as well.

In addition to being larger and less biased, future datasets need more nuanced analysis for benchmarking. All of the publicly released datasets use evaluation metrics that treat every question with equal weight, but some kinds of questions are far easier, either because of bias or because existing algorithms excel at answering that kind of question, e.g., object recognition questions. Some datasets such as COCO-QA have divided VQA questions into distinct categories, e.g., for COCO-QA these are color, counting (number), location, and object. We believe that mean per-question type performance should replace standard accuracy, so each question would not have equal weight in evaluating performance. This would go a long way towards making a VQA algorithm have to perform well at a wide variety of question types to perform well overall, otherwise a system that excelled at answering `Why' questions but was slightly worse than another model at more common questions would not be fairly evaluated. To do this, each question would need to be assigned a category. We believe this effort would make benchmark results significantly more meaningful. The scores on each question type could also be used to compare algorithms to see which kind of questions they excel at.

\section{Conclusions}
VQA is an important basic research problem in computer vision and natural language processing that requires a system to do much more than task specific algorithms, such as object recognition and object detection. An algorithm that can answer arbitrary questions about images would be a milestone in artificial intelligence. We believe that VQA should be a necessary part of any visual Turing test.

In this paper, we critically reviewed existing datasets and algorithms for VQA. We discussed the challenges of evaluating answers generated by algorithms, especially multi-word answers. We described how biases and other problems plague existing datasets. This is a major problem, and the field needs a dataset that evaluates the important characteristics of a VQA algorithm, so that if an algorithm performs well on that dataset then it means it is doing well on VQA in general.

Future work on VQA involves the creation of larger and far more varied datasets. Bias in these datasets will be difficult to overcome, but evaluating different kinds of questions individually in a nuanced manner, rather than using naive accuracy alone, will help significantly. Further work will be needed to develop VQA algorithms that can reason about image content, but these algorithms may lead to significant new areas of research.

\section*{Acknowledgments}
We thank Ronald Kemker for helpful comments on an earlier draft of this paper.

\bibliography{mybibfile}

\end{document}